\documentclass[10pt,twocolumn,letterpaper]{article}

\usepackage[pagenumbers]{cvpr} 

\usepackage{lipsum}

\newcommand\blfootnote[1]{%
  \begingroup
  \renewcommand\thefootnote{}\footnote{#1}%
  \addtocounter{footnote}{-1}%
  \endgroup
}

\newcommand{\benchmark}{NegBench}

\usepackage{graphicx, amsmath, amssymb, caption, subcaption, multirow, overpic, textpos}

\usepackage{tcolorbox}
\usepackage{xcolor}
\usepackage{mdframed}
\newtcbox{\highlightbox}[1][]{nobeforeafter, colframe=gray!50, colback=gray!10, boxrule=0.5pt, arc=4pt, left=1pt, right=1pt, top=1pt, bottom=1pt, #1}
\usepackage{changepage}

\definecolor{customgreen}{HTML}{388E3C}   
\definecolor{customred}{HTML}{D32F2F}     
\definecolor{customblue}{HTML}{1976D2}    

\newlength\savewidth\newcommand\shline{\noalign{\global\savewidth\arrayrulewidth
  \global\arrayrulewidth 1pt}\hline\noalign{\global\arrayrulewidth\savewidth}}
\newcommand{\tablestyle}[2]{\setlength{\tabcolsep}{#1}\renewcommand{\arraystretch}{#2}\centering\footnotesize}

\usepackage{array}
\newcolumntype{x}[1]{>{\centering\arraybackslash}p{#1pt}}

\usepackage{booktabs}
\usepackage{arydshln}

\definecolor{Gray}{gray}{0.5}
\newcommand{\demph}[1]{\textcolor{Gray}{#1}}

\definecolor{PastelGreen}{rgb}{0.13, 0.55, 0.13}%
\newcommand{\emphhigh}[1]{\textcolor{PastelGreen}{#1}}

\definecolor{PastelRed}{rgb}{0.8, 0.13, 0.13}    %

\definecolor{cvprblue}{rgb}{0.21,0.49,0.74}
\usepackage[pagebackref,breaklinks,colorlinks,allcolors=cvprblue]{hyperref}

\def\paperID{2817} 
\def\confName{CVPR}
\def\confYear{2025}

\title{Vision-Language Models Do \emph{Not} Understand Negation}

\author{Kumail Alhamoud$^{1}$
\and
Shaden Alshammari$^{1}$
\and
Yonglong Tian$^{*2}$
\and
Guohao Li$^3$
\and
Philip H.S. Torr$^3$
\and
Yoon Kim$^1$
\and
Marzyeh Ghassemi$^1$\vspace{0.1cm}
\and 
$^1$ MIT \quad $^2$OpenAI \quad $^3$ University of Oxford \\
\href{https://NegBench.github.io}{https://NegBench.github.io}
}

\begin{document}
\maketitle
\begin{abstract}
Many practical vision-language applications require models that understand \emph{negation}, e.g., when using natural language to retrieve images which contain certain objects but not others. Despite advancements in vision-language models (VLMs) through large-scale training, their ability to comprehend negation remains underexplored. This study addresses the question: how well do current VLMs understand negation? We introduce \benchmark{}, a new benchmark designed to evaluate negation understanding across 18 task variations and $79$k examples spanning image, video, and medical datasets. The benchmark consists of two core tasks designed to evaluate negation understanding in diverse multimodal settings: Retrieval with Negation and Multiple Choice Questions with Negated Captions. Our evaluation reveals that modern VLMs struggle significantly with negation, often performing at chance level. To address these shortcomings, we explore a data-centric approach wherein we finetune CLIP models on large-scale synthetic datasets containing millions of negated captions. We show that this approach can result in  a 10\% increase in recall on negated queries and a 28\% boost in accuracy on multiple-choice questions with negated captions. \blfootnote{* Yonglong Tian was at Google Deepmind during this work.}
\end{abstract}    
\vspace{-5pt}
\section{Introduction}
Joint embedding-based Vision-Language Models (VLMs), such as CLIP, have revolutionized how we approach multimodal tasks by learning a shared embedding space where both images and text are mapped together. This shared space enables a variety of applications, including cross-modal retrieval, video retrieval, text-to-image generation, image captioning, and even medical diagnosis~\cite{shenmuch,li2022comprehending,petryk2022guiding,rao2022denseclip,shi2022proposalclip,shridhar2022cliport,baldrati2022effective,li2024cross,sain2023clip,zhang2023biomedclip,lu2024avisionlanguage}. By aligning visual and linguistic representations, these models achieve remarkable performance across domains and are able to model complex interactions between vision and language inputs.

\begin{figure}
    \centering
    \includegraphics[width=1\linewidth, trim={0 0.5cm 0 0}]{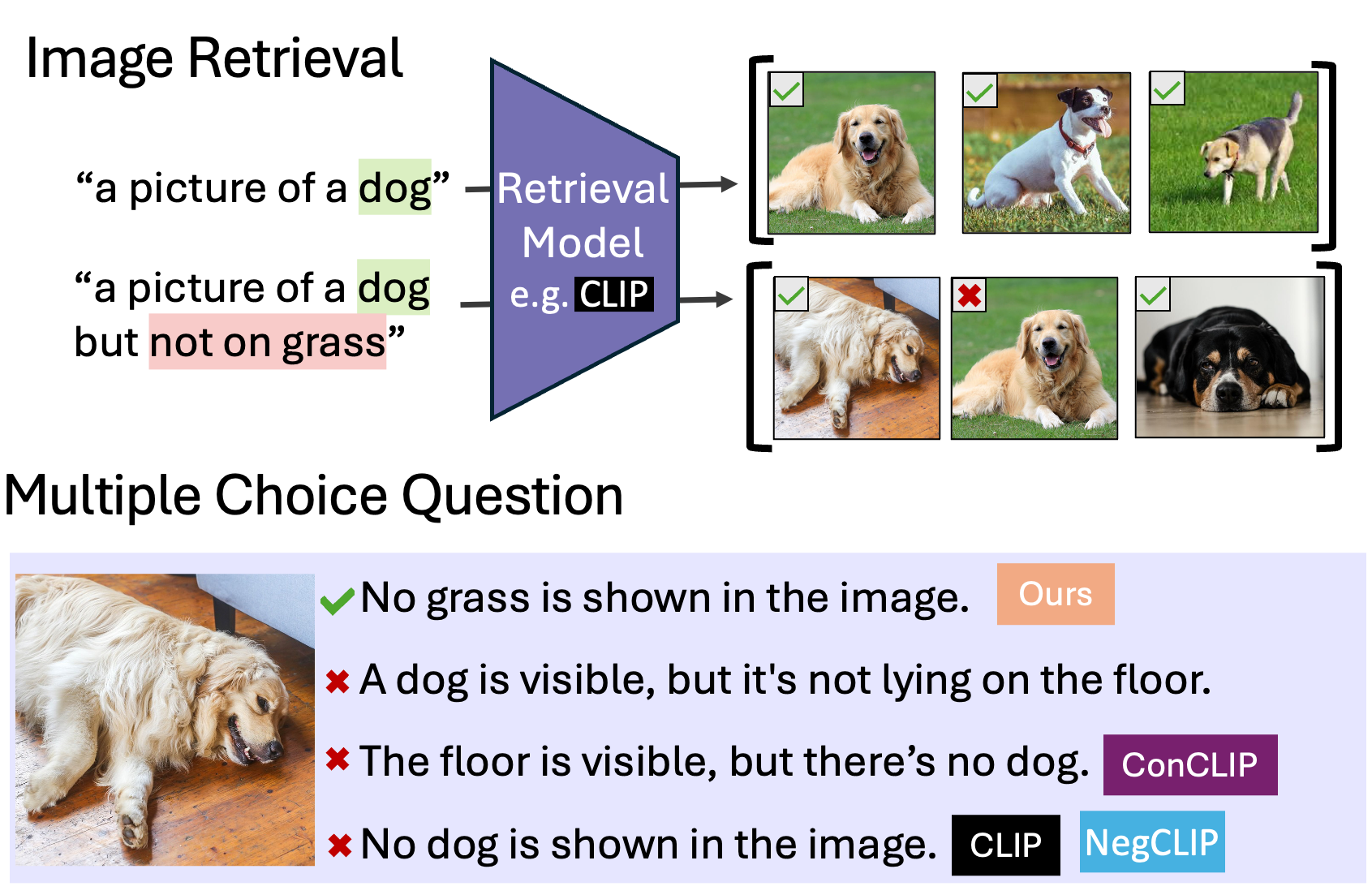}
    \caption{We present \emph{\benchmark{}} with image retrieval and multiple-choice tasks to evaluate negation understanding. CLIP-based models frequently misinterpret negation in both tasks, but we show how a synthetic data approach can improve performance.}
    \vspace{-0.3cm}
    \label{fig:teaser}
\end{figure}

Despite these advances, there is an emerging limitation: these models fail to handle \emph{negation}, which is essential in many real-world scenarios. Negation enables precise communication by specifying what is false or absent~\cite{Horn1989-HORANH, jordan1998power, mukherjee2021effect, morante2021recent}. For example, a radiologist may search for images showing ``bilateral consolidation with no evidence of pneumonia'', or a safety inspector might query ``construction sites with no barriers''. Current benchmarks like CREPE and CC-Neg have introduced limited tests of negation, but they rely on rigid, templated examples that do not reflect the complexity of natural language queries \cite{ma2023crepe, singh2024learn}. As a result, they fall short in evaluating how well VLMs understand negation in practical applications.

To comprehensively evaluate how well VLMs handle negation, we design a multi-level evaluation paradigm inspired by real-world information retrieval systems, where a coarse-grained retrieval step often precedes a fine-grained ranking or selection step~\cite{nogueira2019multi, ma2024fine}.

The first task, Retrieval-Neg, tests whether models can handle real-world queries that mix affirmative and negative statements, such as “a beach with no people” or “a building without windows.” This task challenges the model to retrieve images from diverse datasets based on the presence of certain elements and the absence of others, simulating scenarios found in search engines, content moderation, and recommendation systems. By retrieving several potentially relevant matches (e.g., top-5 retrieval), Retrieval-Neg serves as the coarse-grained retrieval component of our evaluation.

The second task, MCQ-Neg, provides a fine-grained, structured evaluation that directly assesses specific failures in negation. In this task, the model must choose the correct description of an image from several closely related options, where the incorrect choices are hard negatives, differing only by what is affirmed or negated. For instance, in medical diagnostics, consider distinguishing between ``The X-ray shows evidence of pneumonia but no evidence of pleural effusion" and ``The X-ray shows evidence of pleural effusion but no evidence of pneumonia." These statements are linguistically similar but convey opposite diagnoses, requiring the model to parse subtle yet critical differences. 

Through our evaluation pipeline, we uncover a surprising limitation: joint embedding-based VLMs frequently collapse affirmative and negated statements into similar embeddings, treating “a dog” and “no dog” as nearly indistinguishable. This affirmation bias reveals a significant shortcoming that was not sufficiently addressed in previous benchmarks like CREPE or CC-Neg.

Recognizing this critical gap, we then ask: If current models fail to understand negation, can we improve them? To tackle this, we propose a data-centric solution, introducing two large-scale synthetic datasets—CC12M-NegCap and CC12M-NegMCQ—designed to improve negation comprehension. Fine-tuning CLIP-based models on these datasets leads to substantial improvements, including a 10\% increase in recall on negated queries and a 40\% boost in accuracy on multiple-choice questions with negated captions.

The rest of the paper follows a challenge-diagnosis-solution structure. We introduce NegBench to evaluate negation comprehension, analyze VLMs’ affirmation bias, and propose a data-driven solution using synthetic negation examples. We will open-source all models and data to foster research in negation understanding and its applications.

\section{Related Work}
Our work lies within the field of evaluating and advancing foundational vision-language models (VLMs). Joint-embedding models based on CLIP~\cite{radford2021learning} show impressive generalization across visio-linguistic tasks like cross-modal retrieval, image captioning, and visual question answering ~\cite{shenmuch,li2022comprehending,petryk2022guiding,rao2022denseclip,shi2022proposalclip,shridhar2022cliport,baldrati2022effective,li2024cross,sain2023clip} in diverse visual domains, extending beyond natural images to videos and medical images~\cite{narasimhan2021clip,Luo2021CLIP4Clip,Castro_2022_BMVC,zhang2023biomedclip,lu2024avisionlanguage,ikezogwo2024quilt}. We introduce a benchmark and data-centric approach to rigorously evaluate and improve negation understanding in these VLMs.

\vspace{3pt}\noindent\textbf{Negation Understanding in Language and Vision.} 
Recent work showed that large language models perform sub-optimally when tasked with negation understanding \cite{garcia2023dataset, truong2023language}. We go a step further by showing that vision-language models exhibit a more severe affirmation bias, completely failing to differentiate affirmative from negative captions. 

Despite this critical limitation, existing benchmarks provide limited assessments of negation in VLMs. CREPE~\cite{ma2023crepe} and the concurrent work CC-Neg~\cite{singh2024learn} are among the few vision-language benchmarks that include negation, but they focus on compositional understanding and rely on  linguistic templates that fail to reflect the varied ways negation appears in real user queries. In contrast, our proposed benchmark, NegBench, leverages an LLM to generate natural-sounding negated captions, spanning a broader range of negation types and contexts across images, videos, and medical datasets. This systematic design enables a thorough evaluation of VLMs' ability to handle negation in multimodal settings, uncovering unique challenges and failure cases that have not been fully addressed in prior work.

\noindent\textbf{Improving CLIP for Compositionality and Negation.} 
Recent methods have explored improving the generalization abilities of CLIP-like VLMs for visio-linguistic compositionality and limited aspects of negation understanding. For instance, NegCLIP \citep{yuksekgonul2023and} employs composition-aware mining when finetuning CLIP to enhance compositional reasoning, while ConCLIP \citep{singh2024learn} modifies the CLIP loss to incorporate synthetic, template-based negation examples. In the medical domain, negation is a common feature in clinical text reports, often indicating the absence of specific pathologies~\cite{tiu2022expert}. Specialized models like BiomedCLIP~\cite{zhang2023biomedclip} and CONCH~\cite{lu2024avisionlanguage} have been pretrained on millions of biomedical image-text pairs to address a variety of medical tasks, leveraging domain-specific knowledge from large-scale multimodal data. NegBench provides a systematic way to evaluate general-purpose and medical VLMs.

\noindent\textbf{Synthetic Data for Model Training.}
It is common to use synthetic data to improve the performance of models in computer vision \cite{abu2018augmented, chen2019learning,jahanian2022generative,yuan2024realfake}. Recent studies have shown that it is possible to use synthetic data to learn general vision-language representations, with some models trained entirely on synthetic images and captions achieving results comparable to real data \cite{tian2023stablerep, tian2024learning, hammoud2024synthclip}. Our approach is similar in spirit, but it constructs synthetic datasets to teach models a new, complex capability—\emph{negation understanding.}
\begin{figure*}[ht]
\centering
\includegraphics[width=\textwidth]{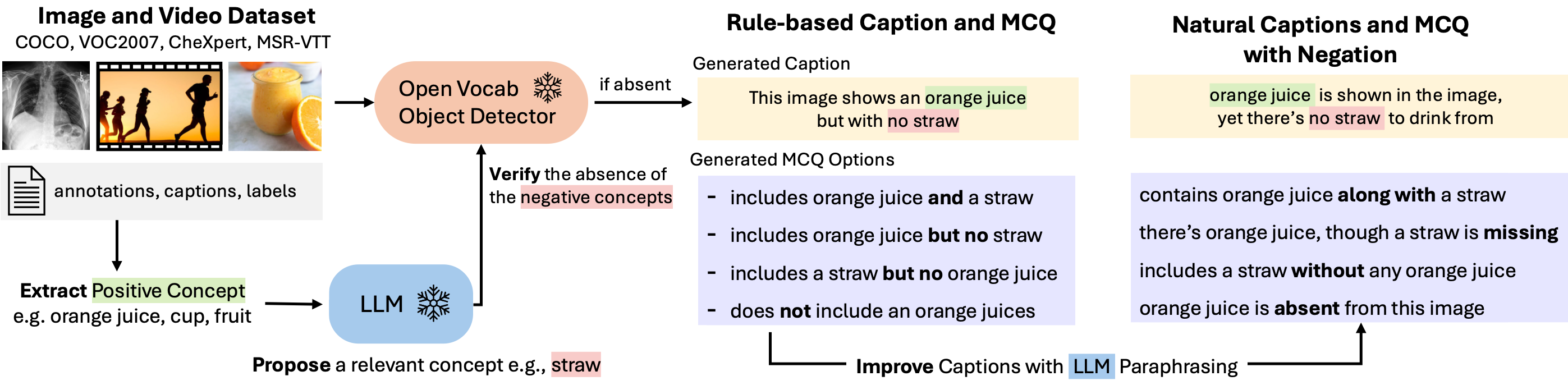}
\caption{\textbf{General Pipeline for Constructing \benchmark{}}. We start by extracting positive concepts from vision datasets. An LLM proposes negative concepts, which are verified with an object detector for datasets without explicit object annotations. We use templates to generate captions with negation, then paraphrase them by an LLM to ensure linguistic variety and robust evaluation of negation understanding.}
\vspace{-12pt}
   \label{fig:sd_pipeline}
\end{figure*}

\section{The Negation Benchmark (\benchmark{})}
\label{sec:benchmark}
We design \benchmark{} as a multi-level evaluation to assess the capacity of joint-based vision-language models to understand negation across different tasks: (1) coarse-grained retrieval, by accurately retrieving images that satisfy specified inclusions and exclusions, and (2) fine-grained question-answering, by selecting the correct description from closely related options, testing the model’s detailed understanding of negation beyond simple retrieval.

In the Retrieval-Neg task, the model retrieves the top-5 images that match both affirmative and negative criteria within a query. In the MCQ-Neg task, the model selects the correct description of an image from options that differ only in the affirmation or negation of specific elements.
\vspace{-5pt}

\subsection{Transforming Datasets for Negation Evaluation}
\label{sec:dataset_transformation}
\paragraph{General Dataset Transformation Overview.}
To implement the two-stage evaluation pipeline of \benchmark{}, we adapt several popular vision datasets, covering images (COCO~\citep{lin2014microsoft}, VOC2007~\cite{everingham2010pascal}), video (MSR-VTT~\citep{xu2016msr}), and specialized medical imaging domains (CheXpert~\citep{irvin2019chexpert}). For each dataset, we identify positive elements $\{pos\}$, which represent objects or concepts present in the image, and negative elements $\{neg\}$, which are absent from the image but commonly associated with the present objects. When available, we use object-level annotations to identify these elements, as in COCO, VOC2007, and CheXpert; for other datasets, we derive positive and negative elements directly from the captions. This flexible approach allows \benchmark{} to extend any vision dataset, whether it includes object-level annotations or captions, to evaluate negation comprehension across diverse tasks and data modalities.

In the Retrieval-Neg task, we modify standard captions by including negations, evaluating how models handle queries that specify both present and absent elements. For example, captions are modified as: ``There is no $x$ in the image. [Original Caption]." or ``[Original Caption]. There is no $x$ in the image." To introduce linguistic diversity, we use LLaMA 3.1~\citep{dubey2024llama} to paraphrase these captions.

For the MCQ-Neg task, we generate multiple-choice questions (MCQs) for each image. The model must identify the correct description based on three linguistic templates: Affirmation, Negation, and Hybrid~\citep{laka1990negation}.

\begin{tcolorbox}[colback=gray!10, boxrule=0pt, sharp corners, colframe=gray!80, left=1mm, right=1mm, top=1mm, bottom=1mm]
\textbf{1. Affirmation}: “This image includes \textcolor{customgreen}{\bf\texttt{A}} (and \textcolor{customgreen}{\bf\texttt{C}}).” 

\textbf{2. Negation}: “This image does not include \textcolor{customred}{\bf\texttt{B}}.” 

\textbf{3. Hybrid}: “This image includes \textcolor{customgreen}{\bf\texttt{A}} but not \textcolor{customred}{\bf\texttt{B}}.”
\end{tcolorbox}

Each MCQ consists of one correct answer and three incorrect answers, which serve as hard negatives, misleading the model if it does not properly understand negation. A correct answer accurately describes the presence of $\{pos\}$ elements or negates $\{neg\}$ elements. A False Affirmation (e.g., ``This image includes $x$" when $x \in \{neg\}$) or a False Negation (e.g., "This image does not include $x$" when $x \in \{pos\}$) highlights the model’s failure to comprehend the image. The Hybrid template further evaluates the model’s ability to combine affirmation and negation in the same caption. These MCQs are also paraphrased using LLaMA 3.1 to increase linguistic diversity.

\subsection{Applicability Across Data Types and Domains} \benchmark{} supports a wide range of data types and domains, enabling comprehensive negation evaluation.

\vspace{2pt}\noindent\textbf{Video Understanding.} Video retrieval tasks introduce temporal complexity, where negation can involve both objects and actions that vary over time. Using MSR-VTT as an example, we prompt LLaMA 3.1~\citep{dubey2024llama} to extract positive and negative elements from each video’s caption. These elements may represent either objects present in the video or actions taking place. For Retrieval-Neg, we create captions specifying both the presence of some elements and the absence of others (e.g., “A person is cooking but not eating”). In MCQ-Neg, we generate multiple-choice questions where the model must select the description that most accurately represents a video segment, requiring it to reason about negation of objects and actions in dynamic scenes.

\vspace{2pt}\noindent\textbf{Medical Image Interpretation with CheXpert.}
Accurate negation understanding is critical in high-stakes domains like medical imaging.  Using the CheXpert dataset~\citep{irvin2019chexpert}, we focus on the most frequent condition \emph{Lung Opacity} and design two binary classification tasks:

\noindent\textit{Task 1: Affirmation Control Task.}
This task evaluates the model’s ability to associate images with specific medical conditions using affirmative statements.

\begin{tcolorbox}[colback=gray!10, boxrule=0pt, sharp corners]\vspace{-7pt}
\textbf{Question}: Which option describes this image?\vspace{5pt}\\
{A}) This image shows Lung Opacity. 

{B}) This image shows Atelectasis.\vspace{-7pt}
\end{tcolorbox}

\noindent\textit{Task 2: Negation Understanding Task.}
This task tests whether the model can correctly interpret negation, distinguishing the presence or absence of a medical condition.

\begin{tcolorbox}[colback=gray!10, boxrule=0pt, sharp corners]\vspace{-7pt}
\textbf{Question}: Which option best describes the image?\vspace{5pt}\\
{A}) This image shows Lung Opacity.

{B}) This image does \textcolor{red}{\emph{not}} show Lung Opacity.\vspace{-7pt}
\end{tcolorbox}

These extensions highlight the adaptability of \benchmark{} to various data types and domains, from general images and videos to specialized medical imaging. This versatility ensures that \benchmark{} provides rigorous, contextually relevant evaluations of negation understanding in VLMs.

\begin{figure}[ht]
    \centering
    \includegraphics[width=0.7\linewidth]{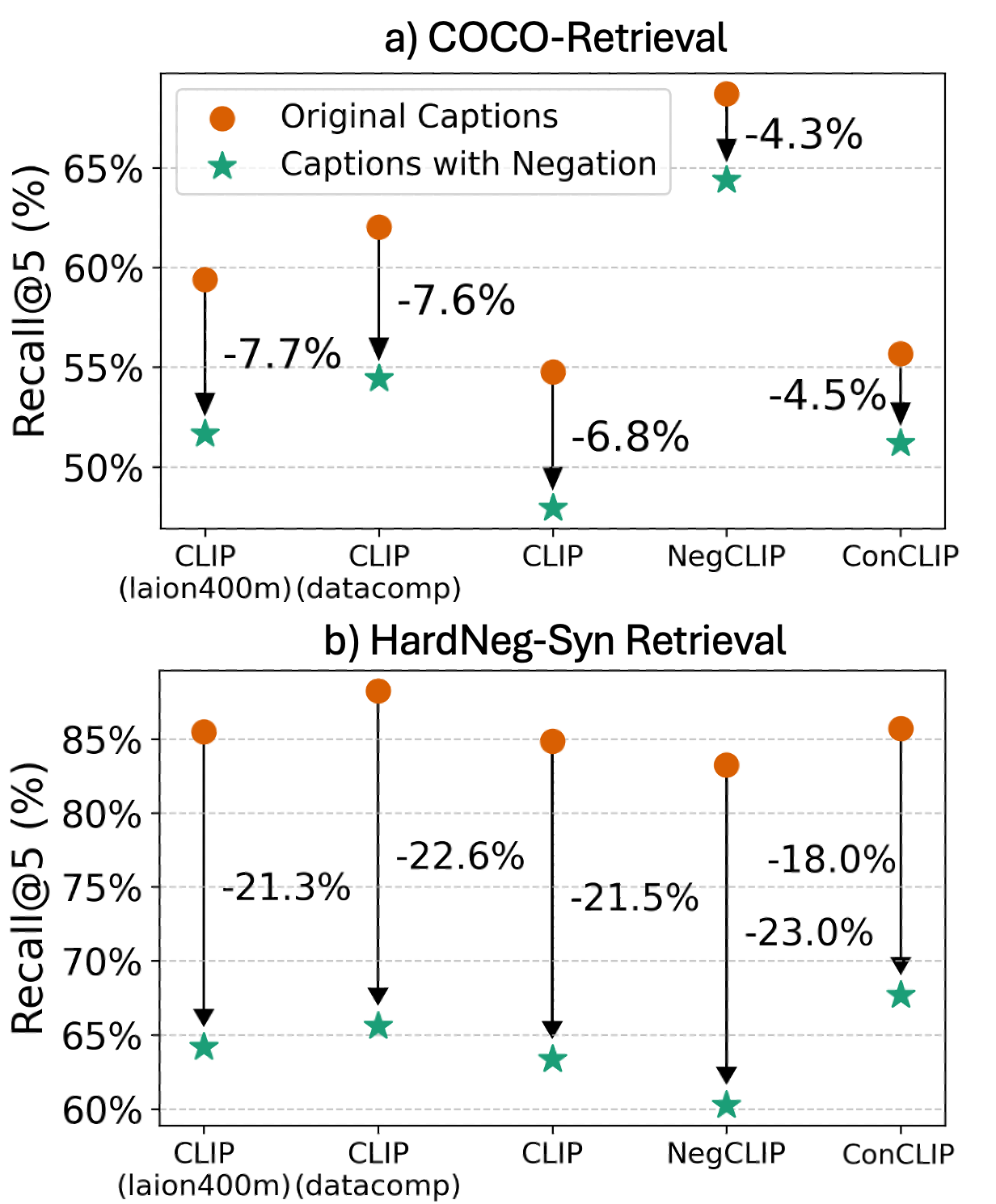}
    \caption{\textbf{Performance drop in recall@5 on (a) COCO and (b) HardNeg-Syn text-to-image retrieval with negated captions (green stars) compared to original captions (orange circles).} All models show substantial drops in performance, with NegCLIP experiencing the largest drop of 23.0\% on HardNeg-Syn, which features hard negatives requiring stronger negation reasoning.}
    \label{fig:ret_drop}
    \vspace{-5pt}
\end{figure}

\begin{figure*}[ht]
\centering
\includegraphics[width=\textwidth]{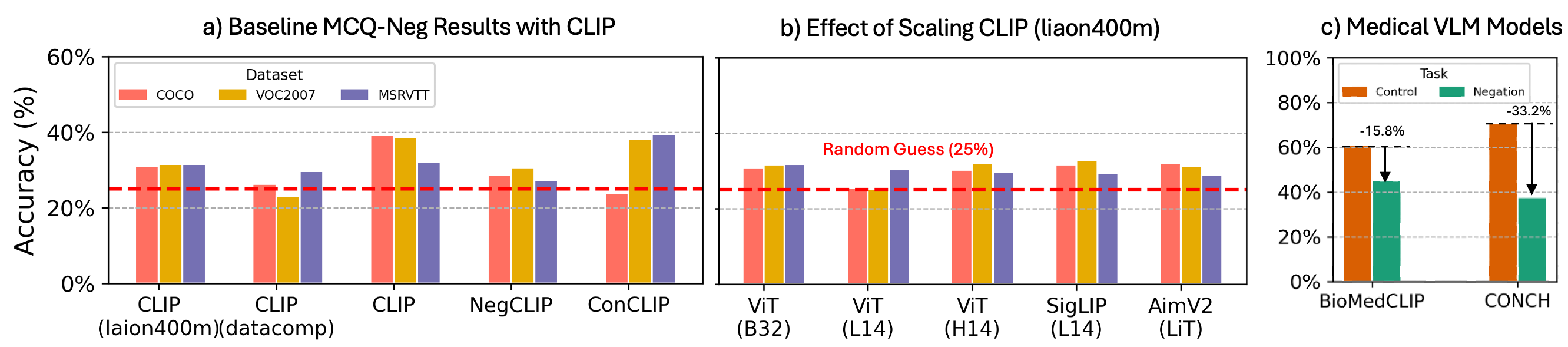} 
\caption{\textbf{MCQ-Neg performance across model families.} \textbf{(a)} CLIP-based models perform near random guessing (shown as a red dashed line), revealing their poor ability to handle negation. \textbf{(b)} Increasing model size (ViT-B$\rightarrow$L$\rightarrow$H) and using more advanced joint-embedding models (SigLIP, AIMV2) does not lead to better negation understanding, despite strong performance on other VLM tasks. \textbf{(c)} Medical VLMs experience large performance drops on negation MCQs, highlighting the risks of affirmation bias in high-stakes applications.}
    \label{fig:mcq_total}
    \vspace{-12pt}
\end{figure*}
\vspace{-10pt}
\subsection{Synthetic Datasets for Controlled Evaluation}
\label{sec:synthetic_data}

To rigorously test negation understanding, we construct \emph{HardNeg-Syn}, a dataset that precisely controls object presence and absence by synthesizing hard negative images.

\vspace{2pt}\noindent\textbf{Motivation and Benefits of Synthetic Data.}
Synthetic data offers several advantages over traditional image datasets. First, by creating “hard negatives”—image pairs that differ only by a single object's presence or absence—we can evaluate the sensitivity of models to negation with minimal confounding variables. Additionally, image datasets like COCO and VOC2007 are limited in the range of visual concepts they cover; COCO has 80 objects while VOC2007 includes only 20. To expand this diversity, we prompt a large language model to propose a broader set of objects, which we use as targets in our synthetic dataset. This approach enables the generation of visually varied scenes that more comprehensively test negation comprehension across a wider array of objects and contexts.

\vspace{2pt}\noindent\textbf{Construction Process for the HardNeg-Syn Evaluation Dataset.}
We create 10,000 image pairs using Stable Diffusion~\citep{Rombach_2022_CVPR}, where each pair includes one image containing a target object and another where it is explicitly absent. To ensure accurate object presence or absence, we use the open-vocabulary object detector OWL-ViT~\citep{minderer2022simple}.

\vspace{-5pt}
\section{\benchmark{} Evaluations: Results and Insights}
In this section, we benchmark the negation abilities of different VLMs using NegBench, comparing models based on their architecture, training data, and training objectives to reveal specific areas where negation understanding remains limited. Specifically, we evaluate five CLIP ViT-B/32 models on Retrieval-Neg and MCQ-Neg tasks. These include OpenAI CLIP~\citep{radford2021learning}, CLIP-laion400m~\citep{schuhmann2021laion}, and CLIP-datacomp~\citep{gadre2023datacomp}, which differ by pretraining dataset, as well as NegCLIP~\citep{yuksekgonul2023and}, trained to improve compositional language understanding, and ConCLIP~\citep{singh2024learn}, trained specifically to improve negation understanding. To handle the video dataset, MSR-VTT, we follow \citep{Castro_2022_BMVC} and encode $4$ uniformly sampled frames per video, averaging their features to obtain the video embedding. For medical tasks, we evaluate CONCH~\citep{lu2024avisionlanguage} and BioMedCLIP~\citep{zhang2023biomedclip}, two medical foundation VLMs. We also assess the impact of scaling up CLIP-laion400m (ViT-B, ViT-L, and ViT-H) to determine if larger embedding model sizes improve negation understanding. In addition, we investigate whether recent joint-embedding models trained with advanced objectives, such as SigLIP (ViT-L)~\citep{zhai2023sigmoid}, or AIMV2~\citep{fini2024multimodal} with Locked-image text Tuning~\citep{zhai2022lit}, offer better performance on negation tasks.

\vspace{2pt}\noindent\textbf{CLIP models struggle with negated queries in retrieval tasks.}
\label{sec:drop}
We evaluate five CLIP-based models on the original COCO text-to-image retrieval task and its Retrieval-Neg version, where captions include negated statements. Across models, performance drops significantly on the negated task. In COCO retrieval (\Cref{fig:ret_drop}a), CLIP-laion400m experiences a 7.7\% drop in recall@5, with CLIP-datacomp and CLIP showing drops of 7.6\% and 6.8\%, respectively. In the more challenging HardNeg-Syn retrieval task (\Cref{fig:ret_drop}b), the performance drops are even more pronounced due to the presence of hard negatives, \ie images that closely resemble positive examples but differ by the exclusion of a single object. Here, NegCLIP, despite its promise for compositional understanding, suffers a 23.0\% drop, while ConCLIP, designed specifically for negation understanding, still declines by 18.0\%. These results suggest that interpreting negation, particularly in the presence of hard negatives, remains a key challenge for retrieval tasks.

\vspace{2pt}\noindent\textbf{MCQ-Neg reveals severe limitations in CLIP models.}
\Cref{fig:mcq_total}a shows that most models perform at or only slightly better than random guessing (indicated by the red dashed line at 25\%) on the MCQ-Neg task. Interestingly, both NegCLIP and ConCLIP fall short from improving over the original OpenAI CLIP NegBench performance. Overall, these results reveal a fundamental limitation of CLIP-like pretraining objectives, which encourage strong associations between visual concepts and specific words, but struggle to interpret nuanced language like negation. Notably, the highest value is CLIP's accuracy on COCO, which is 39\%. However, a score of sub 40\% on a 4-way multiple-choice task is far below an acceptable level, demonstrating that models exhibit a serious lack of negation understanding.

\vspace{2pt}\noindent\textbf{Bigger or newer is not (yet?) better at negation.}
We show in \Cref{fig:mcq_total}b that scaling up the model size from ViT-B/32 (86M parameters) to ViT-L/14 (307M parameters) and ViT-H/14 (632M parameters) does not improve negation understanding. We also evaluate the more recent joint-embedding models SigLIP (ViT-L/14) and AIMV2 (LiT), observing that they too fail to outperform baseline CLIP models on the MCQ-Neg task. Given that AIMV2 represents the state of the art on many vision-language tasks at the time of writing, this further highlights that negation remains a significantly under-addressed challenge in current VLMs.

\vspace{2pt}\noindent\textbf{Critical failures in high-stakes medical tasks.}
\label{sec:mcq_total}
\Cref{fig:mcq_total}c presents the results for the CheXpert MCQ-Neg task, where BioMedCLIP and CONCH exhibit substantial performance drops of 15.8\% and 33.2\%, respectively, when negation is introduced. 
This result is especially concerning in the context of medical diagnostics, where accurate interpretation of negation (e.g., the presence or absence of a condition such as Lung Opacity) is essential for correct diagnoses.

\subsection{Why Do VLMs \emph{Not} Understand Negation?}
\label{sec:mcq_why}

The results from \benchmark{} reveal that CLIP VLMs struggle with different forms of negation understanding, motivating a deeper analysis into the underlying causes of these failures. In this section, we examine model performance across different MCQ types and analyze the embedding spaces of various models to uncover specific shortcut strategies that limit their negation comprehension.

\begin{figure*}[h]
\centering
\includegraphics[width=\textwidth]{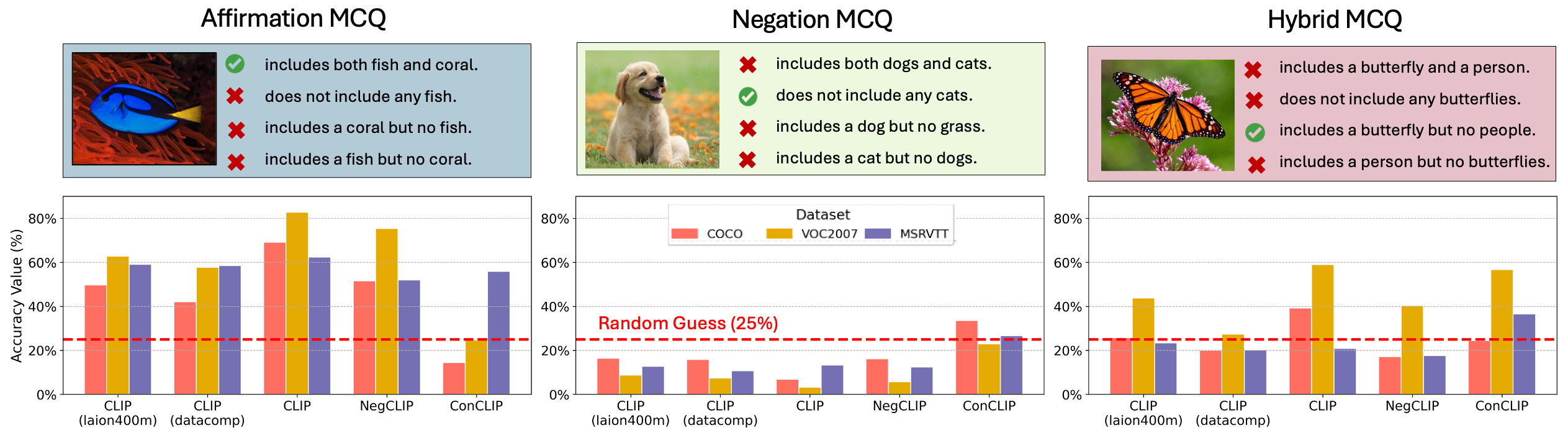}
\caption{\textbf{Performance by MCQ type: Affirmation, Negation, and Hybrid.} CLIP-like models exhibit strong \emph{affirmation bias}—they perform well on Affirmation MCQs (left panel), but fail on Negation MCQs (middle panel), often performing much below random chance.}
    \label{fig:mcq_breakdown}
\vspace{-11pt}
\end{figure*}

\vspace{2pt}\noindent\textbf{Model performance varies widely across MCQ types.}
To understand why models struggle to perform better than random chance, we categorize the MCQs into three types based on the correct answer template: Affirmation, Negation, and Hybrid. \Cref{fig:mcq_breakdown} compares model accuracy across these MCQ types. All models perform poorly on Negation MCQs, reflecting a general struggle with negation understanding (middle panel). In contrast, performance on Affirmation MCQs is substantially higher (left panel)—for instance, CLIP achieves 82\% accuracy on Affirmation MCQs for VOC2007, but only 3\% on Negation MCQs, revealing a severe affirmation bias in all models (except ConCLIP.)

To understand this behavior, we analyze the types of sentences models tend to select when making mistakes. Most models frequently choose Negation sentences that incorrectly negate existing objects (see template selection frequencies in the appendix). This likely stems from the task design: 67\% of MCQs (Negation and Hybrid) do not contain a correct Affirmative option, which causes biased models to default to statements like ``This image does not include \{pos\}." These results suggest that models trained with CLIP-like objectives often adopt shortcut strategies that ignore critical words such as ``no." We refer to this tendency as the \emph{affirmation bias} of CLIP-like models.

While ConCLIP appears less susceptible to affirmation bias, it does not outperform other models in NegBench, as its accuracy on Negation and Hybrid MCQs remains low. As we will show next, ConCLIP suffers from a different kind of bias that hinders its usability: it maps templated Hybrid captions to the same location in its embedding space.

\begin{figure*}[ht]
    \centering
    \begin{subfigure}{0.95\textwidth}
        \centering
        \includegraphics[width=\linewidth]{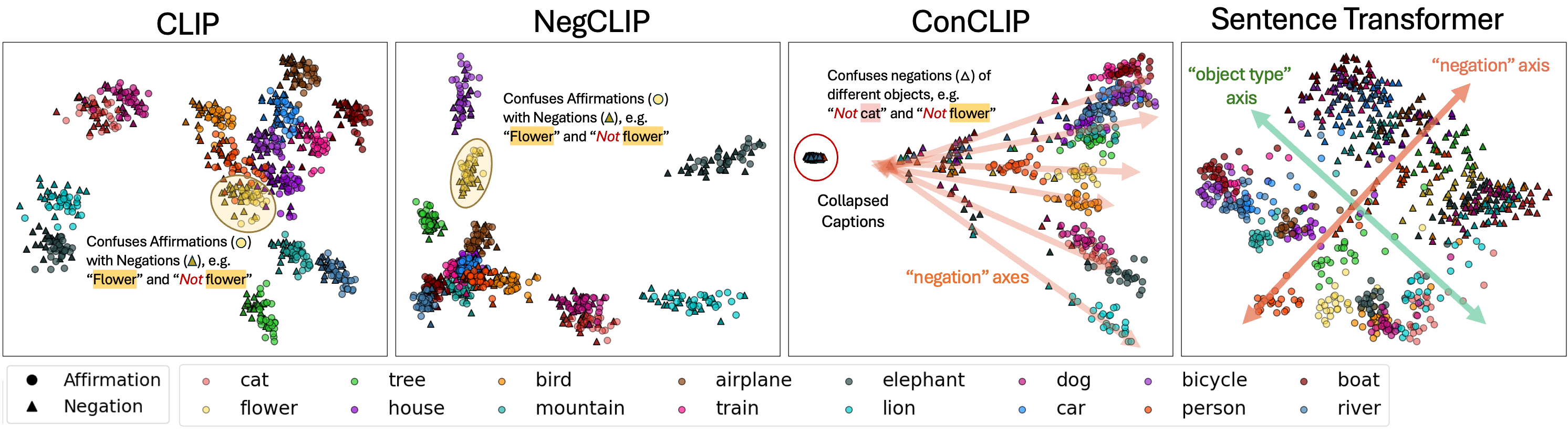}
        \caption{PCA embeddings for affirmative (dots) and negated (triangles) captions. }
        \label{fig:caption_projection_pca}
    \end{subfigure}
    
    \vspace{-0.1pt} 

    \begin{subfigure}{0.95\textwidth}
        \centering
        \includegraphics[width=\linewidth]{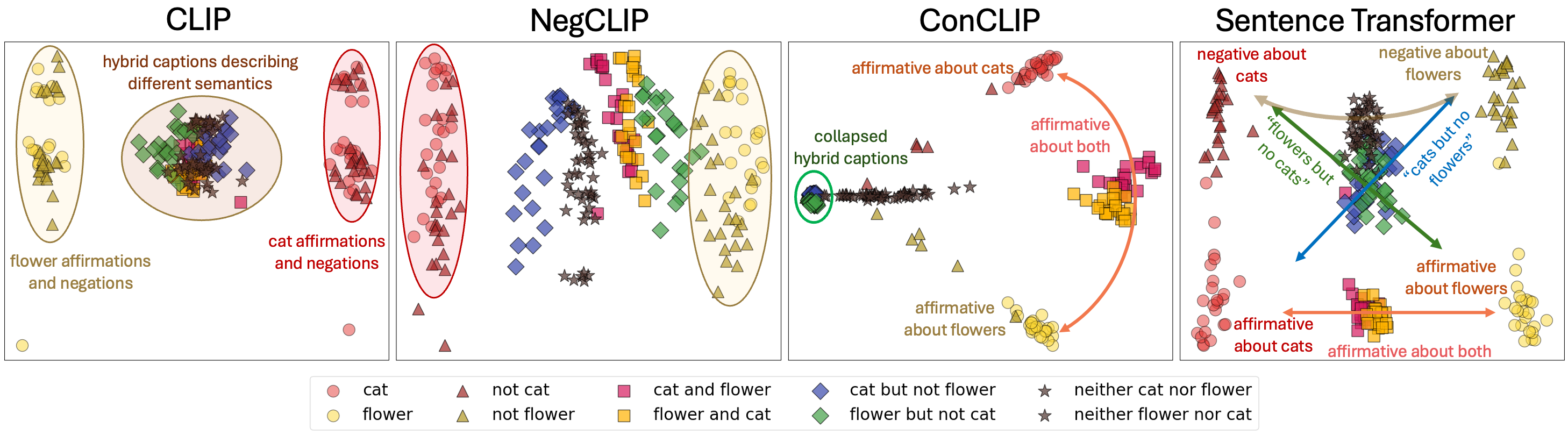}
        \caption{PCA embeddings for hybrid captions (diamonds) and cases where two objects are negated (stars) or affirmed (squares).}
        \label{fig:hybrid_projection_pca}
    \end{subfigure}

    \caption{\textbf{PCA Projections of Caption Embeddings Across Models.} CLIP and NegCLIP lack separation between affirmative and negated captions. ConCLIP treats all negated captions as identical, regardless of the object type, while the Sentence Transformer shows more ideal separability along both 'object type' and 'negation' dimensions.}
    \label{fig:combined_pca}
    \vspace{-12pt}
\end{figure*}

\noindent\textbf{Embedding analysis reveals VLM shortcut strategies.}
To investigate potential shortcut strategies, we analyze the embedding spaces of various models using 24 Affirmative (``X") and 24 Negated (``Not X") templates to create 48 captions per object. We apply PCA to the resulting embeddings (\Cref{fig:caption_projection_pca}). The templates are detailed in the appendix. 

We observe varying behaviors across models. The overlapping embeddings for affirmative and negated captions in \emph{CLIP} and \emph{NegCLIP} suggest that these models do not distinguish between positive and negative statements, possibly due to a ``bag-of-words" shortcut strategy~\citep{geirhos2020shortcut, yuksekgonul2023and} that overlooks negation words. This explains why both models incorrectly select the Negation template, which negates positive objects, in \Cref{fig:mcq_breakdown}. \emph{CoNCLIP} separates positive and negative captions but fails to distinguish between negative captions of different objects, collapsing all negative caption embeddings toward a single point (red circle). 

We include the embeddings of a text-only Sentence Transformer~\citep{reimers-2019-sentence-bert} as a reference that effectively differentiates affirmative and negated captions along distinct ``object type" and ``negation" axes, exemplifying ideal separation.
\newpage
\noindent\textbf{Hybrid captions reveal more evidence of collapsed embeddings.}
\Cref{fig:hybrid_projection_pca} extends the previous analysis to hybrid captions that combine affirmations and negations. It provides further evidence that \emph{ConCLIP} employs a shortcut strategy for embedding linguistic negation, with hybrid and negated captions collapsing towards a single point (green circle), indicating significant compression along the negation axis. While \emph{CLIP} and \emph{NegCLIP} struggle to distinguish affirmative from negative statements, \emph{NegCLIP} shows better separation for hybrid captions, which appear collapsed in the CLIP embedding space. This suggests that NegCLIP's poor performance on Hybrid MCQs might be due to a misalignment between the text and image encoders, rather than an inability to understand hybrid sentence structure. 
In contrast, the \emph{Sentence Transformer} effectively distinguishes between different caption types and provides semantically guided representations. For example, it aligns ``flowers but not cats" along the line connecting ``flowers" and ``not cats."
\section{A Data-Centric Approach for Improving Negation Understanding}
\label{sec:data_centric_solution}
We hypothesize that the tendency of CLIP-based models to rely on linguistic shortcuts, which hinders their negation understanding as explored in \Cref{sec:mcq_why}, stems from training data limitations.
In CLIP, training data lacks examples with explicit negation, leaving it unable to distinguish negated and affirmed concepts. In contrast, ConCLIP's training data overfits to a single hybrid linguistic template, limiting its ability to generalize across varied negation structures. Next, we explore data-centric strategies to address these gaps, introducing a dataset that includes diverse negation examples spanning a range of linguistic styles.

\subsection{Synthesizing a Fine-Tuning Negation Dataset}
We augment the CC12M dataset~\citep{changpinyo2021cc12m}, which contains approximately 10 million image-text pairs, to generate two synthetic datasets with negation: {CC12M-NegCap} and {CC12M-NegMCQ}. Our goal is to expose models to a wide variety of negation scenarios and improve their ability to encode negated statements. The process follows these steps:
\begin{enumerate}
    \item \textbf{Object Extraction}: Using LLaMA 3.1~\citep{dubey2024llama}, we extract positive objects (those mentioned in the caption) and negative objects (contextually relevant but not present) from each image-caption pair in CC12M.

    \item \textbf{Visual Verification}: An open-vocabulary object detector~\citep{minderer2022simple} verifies the presence of positive objects and ensures the absence of the negative objects in the image. This step is crucial to avoid introducing incorrect negations that could confuse the model.

    \item \textbf{Caption Generation}: For each image, we generate multiple new captions that incorporate negated objects into the original captions. LLaMA 3.1 is used to ensure the generated captions are natural-sounding and reflect realistic negation scenarios found in retrieval queries.
\end{enumerate}
\vspace{3pt}
We construct two variants of the synthetic dataset.\textbf{CC12M-NegCap} includes three captions per image with incorporated negated objects, totaling approximately 30 million captions. \textbf{CC12M-NegMCQ} includes four captions per image: one correct and three hard negatives based on object annotations, offering stronger training signals for fine-grained negation understanding and resulting in around 40 million captions. To balance broad retrieval with fine-grained negation capabilities, we introduce \textbf{CC12M-NegFull}, a comprehensive dataset that combines CC12M-NegCap and CC12M-NegMCQ. We will release the extracted object annotations for each image in CC12M, along with the corresponding URLs, and all the generated captions in CC12M-NegFull. This will help the community build on our dataset and advance research in negation understanding and multimodal retrieval.

\subsection{Fine-Tuning with Negation-Enriched Data}
\vspace{2pt}\noindent\textbf{Standard CLIP Objective on CC12M-NegCap.}
Let \( \mathcal{B}_{\text{cap}} = \{(I_i, T_i)\}_{i=1}^N \) represent a batch of \( N \) image-caption pairs from CC12M-NegCap, where each image \( I_i \) is paired with a caption \( T_i \) that describes present and absent objects in the image. For each batch \( \mathcal{B}_{\text{cap}} \), we compute a similarity matrix \( S \in \mathbb{R}^{N \times N} \), where each element \( S_{j,k} \) represents the cosine similarity between the \( j \)-th image and the \( k \)-th caption. The CLIP objective applies a symmetric cross-entropy loss over this matrix, encouraging high similarity for correct image-caption pairs and low similarity for incorrect pairs. This loss is denoted as \( \mathcal{L}_{\text{CLIP}}(\mathcal{B}_{\text{cap}}) \) and provides the model with diverse negation examples in a contrastive learning setup.

\vspace{2pt}\noindent\textbf{Multiple-Choice Objective on CC12M-NegMCQ.}

Let \( \mathcal{B}_{\text{mcq}} = \{(I_i, \{T_{i,1}, \dots, T_{i,C}\})\}_{i=1}^M \) be a batch of \( M \) examples from CC12M-NegMCQ, where each image \( I_i \) is paired with \( C \) captions \( \{T_{i,j}\}_{j=1}^C \). One caption correctly describes the image, while the others serve as hard negatives. For our experiments, we set \( C = 4 \). To fine-tune on CC12M-NegMCQ, we compute the cosine similarity between each image and its four caption options, generating a set of logits for each image-option pair.

The multiple-choice loss \( \mathcal{L}_{\text{MCQ}}(\mathcal{B}_{\text{mcq}}) \) is then computed by applying a cross-entropy loss over the logits, with the correct answer index as the target. This loss encourages the model to assign higher similarity to the correct caption and lower similarity to the hard negative captions:
\begin{equation}
\mathcal{L}_{\text{MCQ}}(\mathcal{B}_{\text{mcq}}) = - \frac{1}{M} \sum_{i=1}^M \log \frac{\exp(\text{logits}_{i, c_i})}{\sum_{j=1}^{C} \exp(\text{logits}_{i, j})},
\label{eq:mcq_loss}
\end{equation}
where \(c_i\) indicates the index of the correct caption describing the \(i\)-th image.

\noindent\textbf{Combined Training Objective.}
The final objective combines the contrastive loss on CC12M-NegCap with the MCQ loss on CC12M-NegMCQ, weighted by \(\alpha\) to balance their contributions. The total loss for one batch is:
\begin{equation}
\mathcal{L}_{\text{Total}} = \alpha \mathcal{L}_{\text{CLIP}}(\mathcal{B}_{\text{cap}}) + (1-\alpha) \mathcal{L}_{\text{MCQ}}(\mathcal{B}_{\text{mcq}}).
\label{eq:total_loss}
\end{equation}

\noindent\textbf{Evaluation Protocol.} To assess the impact of our data-centric approach, we fine-tune two pretrained models (OpenAI CLIP and NegCLIP) on CC12M-NegCap using the contrastive loss \( \mathcal{L}_{\text{CLIP}} \). Additionally, we fine-tune both models on the combined CC12M-NegCap and CC12M-NegMCQ datasets using \( \mathcal{L}_{\text{Total}} \) in \Cref{eq:total_loss}. For comparison, we fine-tune these models on the original CC12M dataset to isolate the effect of our negation-enriched datasets. Our goal is to demonstrate that CLIP models can significantly improve their understanding of negation with the right data. 

We evaluate the models on two tasks: (i) text-to-image and text-to-video retrieval on COCO and MSR-VTT, both with and without negated queries, and (ii) image-to-text and video-to-text MCQ tasks, where models select the correct caption from four options. The results are shown in \Cref{tab:performance}.

\vspace{2pt}\noindent\textbf{Results.} Fine-tuning CLIP and NegCLIP on negation-enriched data leads to consistent and substantial improvements across both retrieval and MCQ tasks. On COCO, fine-tuning CLIP with CC12M-NegCap improves R-Neg@5 from 48.0\% to 57.8\%, while MCQ accuracy rises from 39.2\% to 47.3\% (\textbf{+8.1}). Similarly, NegCLIP's MCQ score improves from 28.6\% to 40.4\% (\textbf{+11.8}) with the same data. Larger MCQ gains are observed when training with CC12M-NegFull, which includes both contrastive and MCQ supervision: CLIP and NegCLIP achieve MCQ accuracies of 54.4\% and 56.2\%, respectively, corresponding to relative gains of \textbf{+15.2} and \textbf{+27.6} over their initial baselines. Similar trends also hold on the video dataset MSR-VTT. These results demonstrate that leveraging our high-quality synthetic dataset can effectively enhance VLM negation understanding.

\vspace{7pt}
\begin{table}[h]
\small
\centering

\subfloat[COCO Evaluation]{
\tablestyle{1pt}{1.0}
\begin{tabular}{x{35}x{54}|x{37}x{46}x{40}c}
\textbf{Model} & \textbf{Fine-tune data} &
\textbf{R@5 ($\uparrow$)} &
\textbf{R-Neg@5 ($\uparrow$)} &
\textbf{MCQ ($\uparrow$)} & \\ 
\shline
\multirow{4}{*}{CLIP} & \demph{None} &
\demph{54.8} & \demph{48.0} & \demph{39.2} & \\ 
& CC12M &
58.8 & 54.5 & 34.7 ($\downarrow$4.5) & \\ 
& \textbf{{\scriptsize CC12M-}NegCap} &
\textbf{58.5} & \textbf{57.8} & \textbf{47.3} (\emphhigh{$\uparrow$8.1}) & \\ 
& \textbf{{\scriptsize CC12M-}NegFull} &
\textbf{54.2} & \textbf{51.9} & \textbf{54.4} (\emphhigh{$\uparrow$15.2}) & \\ 
\hline
\multirow{4}{*}{NegCLIP} & \demph{None} &
\demph{68.7} & \demph{64.4} & \demph{28.6} & \\ 
& CC12M &
70.2 & 66.0 & 28.9 ($\uparrow$0.3) & \\ 
& \textbf{{\scriptsize CC12M-}NegCap} &
\textbf{68.6} & \textbf{67.5} & \textbf{40.4} (\emphhigh{$\uparrow$11.8}) & \\ 
& \textbf{{\scriptsize CC12M-}NegFull} &
\textbf{69.0} & \textbf{67.0} & \textbf{56.2} (\emphhigh{$\uparrow$27.6}) & \\ 
\end{tabular}	
} 
\\
\vspace{.5em}
\subfloat[MSR-VTT Evaluation]{
\tablestyle{1pt}{1.0}
\begin{tabular}{x{35}x{54}|x{37}x{46}x{40}c}
\textbf{Model} & \textbf{Fine-tune data} &
\textbf{R@5 ($\uparrow$)} &
\textbf{R-Neg@5 ($\uparrow$)} &
\textbf{MCQ ($\uparrow$)} & \\ 
\shline
\multirow{4}{*}{CLIP} & \demph{None} &
\demph{50.6} & \demph{45.8} & \demph{32.1} & \\ 
& CC12M &
53.7 & 49.9 & 30.8 ($\downarrow$1.3) & \\ 
& \textbf{{\scriptsize CC12M-}NegCap} &
\textbf{54.1} & \textbf{53.5} & \textbf{41.5} (\emphhigh{$\uparrow$9.4}) & \\ 
& \textbf{{\scriptsize CC12M-}NegFull} &
\textbf{46.9} & \textbf{43.9} & \textbf{44.9} (\emphhigh{$\uparrow$12.8}) & \\ 
\hline
\multirow{4}{*}{NegCLIP} & \demph{None} &
\demph{53.7} & \demph{51.0} & \demph{27.3} & \\ 
& CC12M &
56.4 & 52.6 & 31.6 ({$\uparrow$4.3}) & \\ 
& \textbf{{\scriptsize CC12M-}NegCap} &
\textbf{56.5} & \textbf{54.6} & \textbf{39.8} (\emphhigh{$\uparrow$12.5}) & \\ 
& \textbf{{\scriptsize CC12M-}NegFull} &
\textbf{54} & \textbf{51.5} & \textbf{46.2} (\emphhigh{$\uparrow$18.9}) & \\ 
\end{tabular}	
} 

\caption{\textbf{Comparison of fine-tuning datasets} on performance metrics across COCO and MSR-VTT, fine-tuned on respective datasets and evaluated on retrieval and MCQs. Differences in MCQ accuracy from the baseline are shown, with increases of \emphhigh{+8} or more highlighted. Fine-tuning on negation-enriched data significantly improves negation understanding (R-Neg and MCQ).}
\label{tab:performance}
\vspace{-10pt}
\end{table}

\noindent\textbf{Ablation: Effect of varying $\alpha$.} The table below shows the impact of varying the weight factor $\alpha$ in the combined loss $\mathcal{L}_{\text{Total}} = \alpha \mathcal{L}_{\text{CLIP}} + (1 - \alpha ) \mathcal{L}_{\text{MCQ}}$ when fine-tuning CLIP on CC12M-NegFull. As $\alpha$ increases, more weight is placed on the original CLIP contrastive objective, while a lower $\alpha$ emphasizes the MCQ loss. Properly tuning $\alpha$ is important to balance between fine-grained MCQ and standard retrieval.
\begin{center}
\tablestyle{2pt}{1.2}	
\begin{tabular}{c|x{28}x{28}x{28}x{28}x{28}}
$\alpha$ & 0 & 0.5 & 0.9 & 0.99 & 1 \\
\shline
COCO Recall@5 (\%) & 33.9 & 37.3 & 47.6 & 54.2 & 58.5 \\
COCO MCQ Acc (\%) & 59.4 & 53.7 & 54.6 & 54.4 & 47.3 \\
\end{tabular}
\vspace{0em}
\end{center}
\vspace{-6pt}
\section{Discussion and Conclusions}
\vspace{2pt}\noindent\textbf{Implications.} Our findings point to two broader implications for enhancing language understanding in VLMs. From a data perspective, pretraining datasets should include a diverse array of language constructs, especially those involving nuanced expressions like negation or complex syntactic structures, to help models capture the subtleties of human language. Currently, many VLMs are pretrained on datasets that primarily consist of straightforward, affirmative statements, which might limit the models' ability to understand more subtle language elements. From a learning perspective, our results suggest that a combination of contrastive learning and MCQ supervised training can improve coarse-grained retrieval and fine-grained negation understanding. We experimented with different values of $\alpha$ in \Cref{eq:total_loss}, which revealed a tradeoff in performance. This suggests that alternative or supplementary training objectives beyond contrastive learning could enhance models' sensitivity to nuanced language, enabling more robust applications in real-world settings where precise language interpretation is~essential. 

\vspace{2pt}\noindent\textbf{Summary.} This paper introduces \emph{NegBench} to systematically evaluate negation understanding in VLMs. Our findings reveal that CLIP-based models exhibit a strong affirmation bias, limiting their application in scenarios where negation is critical, such as medical diagnostics and safety monitoring. Through synthetic negation data, we offer a promising path toward more reliable models. While our synthetic data approach improves negation understanding, challenges remain, particularly with fine-grained negation differences.
\newpage
\noindent\textbf{Acknowledgments.}
This work was supported in part by a National Science Foundation (NSF) 22-586 Faculty Early Career Development Award (\#2339381), a Gordon \& Betty Moore Foundation award, a Google Research Scholar award, and a UKRI
grant Turing AI Fellowship (EP/W002981/1). The authors would like to thank Walter Gerych, Olawale Salaudeen, and Mark Hamilton for valuable discussions and feedback.

{
    \small
    \bibliographystyle{ieeenat_fullname}
    \bibliography{main}
}

\clearpage
\appendix
\section*{Appendix}
\label{sec:supplementary}

\section{Evaluating LLaVA on NegBench MCQs}
\begin{figure}[ht]
    \centering
    \includegraphics[width=1.\linewidth]{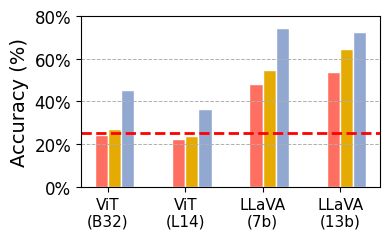}
    \caption{\textbf{Caption.}}
    \label{fig:llava}
\end{figure}

In the main paper, we proposed a novel evaluation paradigm for negation understanding, aimed at simulating real-world scenarios as closely as possible. We then proceeded to evaluate joint embedding-based VLMs, particularly CLIP models, which are the dominant models for multimodal retrieval tasks, in addition to being popular for text-to-image generation, image captioning, and medical multimodal tasks. 
However, we recognize that there are other VLMs that can be useful in certain settings. In particular, instruction-tuned VLMs like LLaVA open up the path for conversational VLM chatbots. In this section, we evaluate LLaVA on the three natural image MCQ tasks in NegBench (COCO, VOC2007, and HardNeg-Syn). The results are in \Cref{fig:llava}. 

\paragraph{LLaVA, an instruction-tuned VLM, demonstrates improvement.}
\Cref{fig:llava} shows that LLaVA significantly outperforms CLIP models on the MCQ-Neg tasks. This is particularly notable because LLaVA uses a CLIP ViT-L/14 vision encoder, which we have shown in \Cref{fig:mcq_total} to struggle with negation. The key advantage of LLaVA might be in its use of the Vicuna LLM for text encoding. Unlike CLIP, which is pretrained on vision-language pairs that predominantly contain affirmative image captions, LLMs like Vicuna are trained on diverse textual corpora that include both affirmations and negations. This broader exposure allows LLaVA to better interpret negated statements. Additionally, LLaVA uses a learned projection layer to align vision and language representations, in contrast to CLIP’s contrastive learning objective, which tends to ignore word order and subtle linguistic cues like negation~\cite{yuksekgonul2023and}. We further explore these differences in \Cref{fig:template_frequency}.

\paragraph{Limitations of LLaVA as a retrieval system.}
While LLaVA demonstrates improved negation understanding, it has significant limitations as a retrieval model compared to CLIP. CLIP learns a joint image-text embedding space, making it highly efficient for retrieval tasks by simply embedding both images and texts, and then computing cosine similarities. In contrast, LLaVA processes a single image-text pair at a time and generates text output, which makes image-to-text retrieval feasible only if all possible captions can fit into the model’s context window. For MCQ-Neg, we applied this method by presenting the image alongside all possible captions and prompting LLaVA to select the correct one. However, this approach does not scale well with a large number of candidates and is not applicable for text-to-image retrieval, where fitting all dataset images into the context window is impractical. Therefore, advancing models like CLIP is crucial for real-world multimodal retrieval with negation. In the paper, we explored the data-centric reasons behind CLIP’s failures in negation understanding and proposed synthetic data strategies to address them.

\section{A Closer Look at VLM Negation Failures}
To better understand the negation failures of VLMs, we further analyze the models' tendency to select specific template types when answering multiple-choice questions (MCQs) and provide further analysis into the embedding space of these models.

\subsection{Template Selection Frequency}
\begin{figure}[h]
\centering
\includegraphics[width=\linewidth]{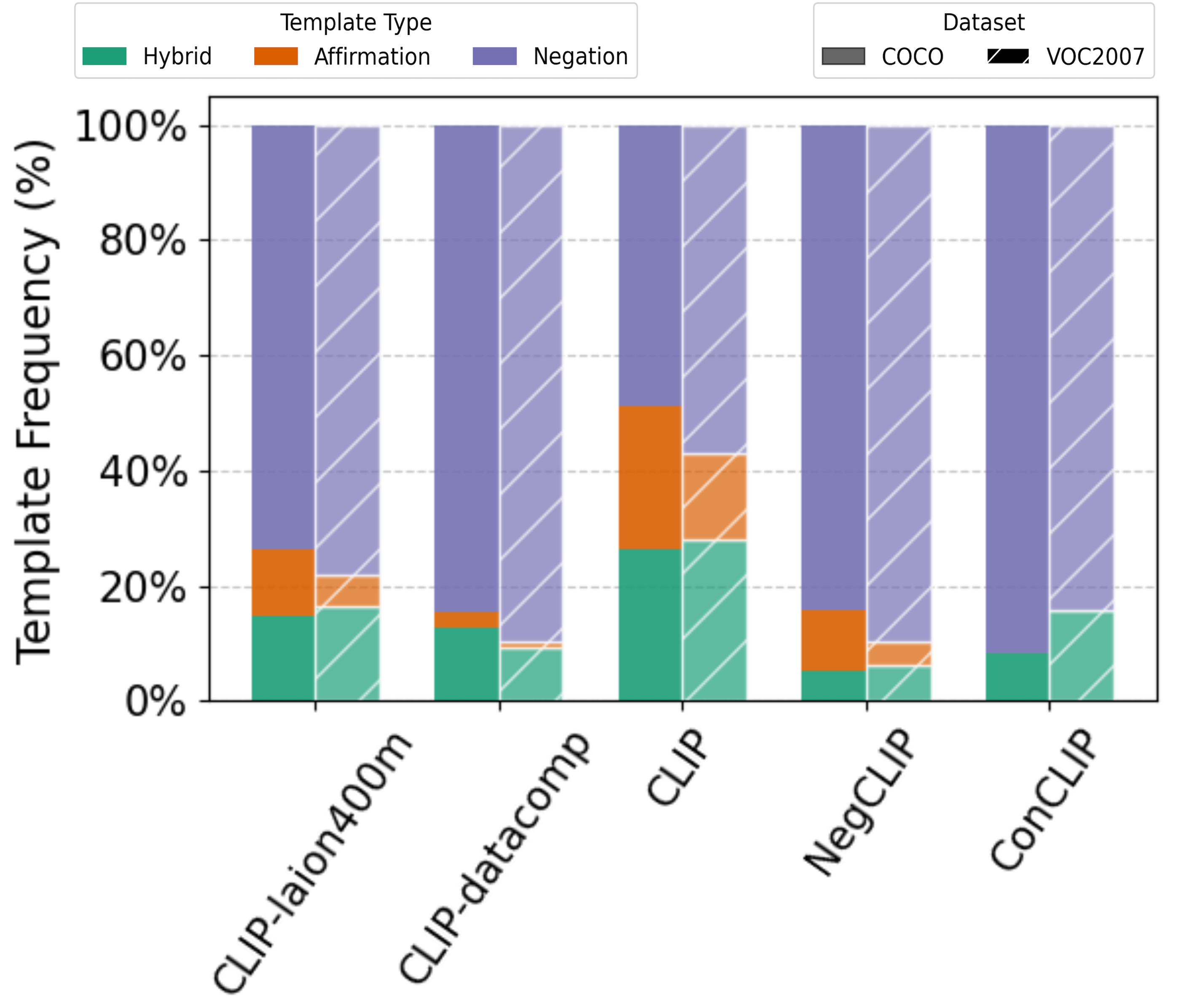}
\caption{Template selection frequency for various models on COCO and VOC2007 datasets, broken down by template type (Affirmation, Negation, Hybrid).}
    \label{fig:template_frequency}
\end{figure}

\Cref{fig:template_frequency} analyzes the frequency with which different models select specific template types (Affirmation, Negation, Hybrid) when answering multiple-choice questions, regardless of the correct answer. This analysis helps to reveal potential biases in model behavior and understand why models may struggle with negation. As shown in \Cref{fig:mcq_breakdown} from the paper, most models perform poorly on Negation MCQs, reflecting a general struggle with negation understanding.

\begin{figure*}[h]
\centering
\includegraphics[width=\textwidth]{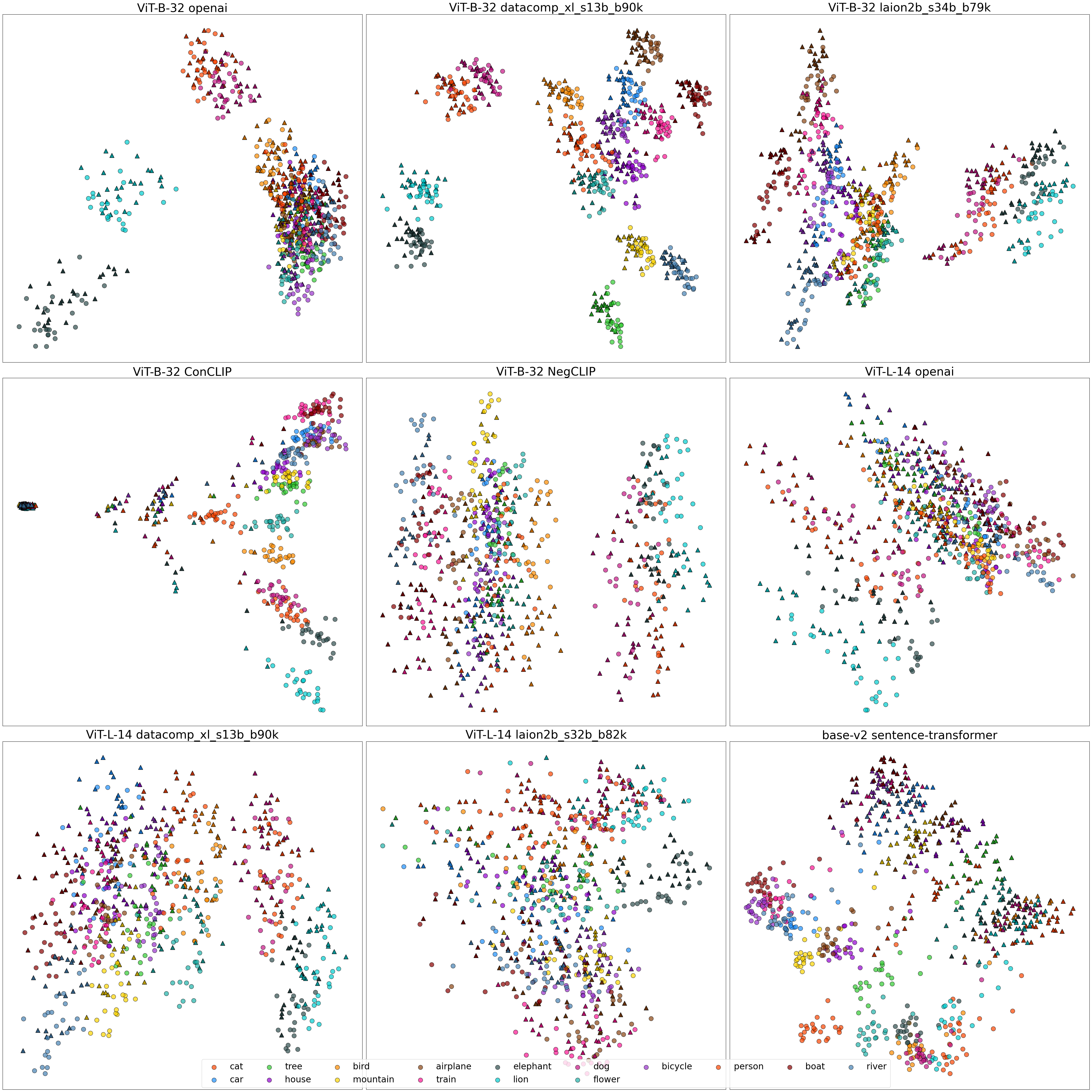}
\caption{PCA projections of caption embeddings for various CLIP models and the Sentence Transformer. Each point represents a caption embedding. This figure complements \Cref{fig:combined_pca} by providing a broader view of embedding separation across different VLMs.}
    \label{fig:all_embeddings}
\end{figure*}

\begin{figure}[h]
\centering
\includegraphics[width=0.5\textwidth]{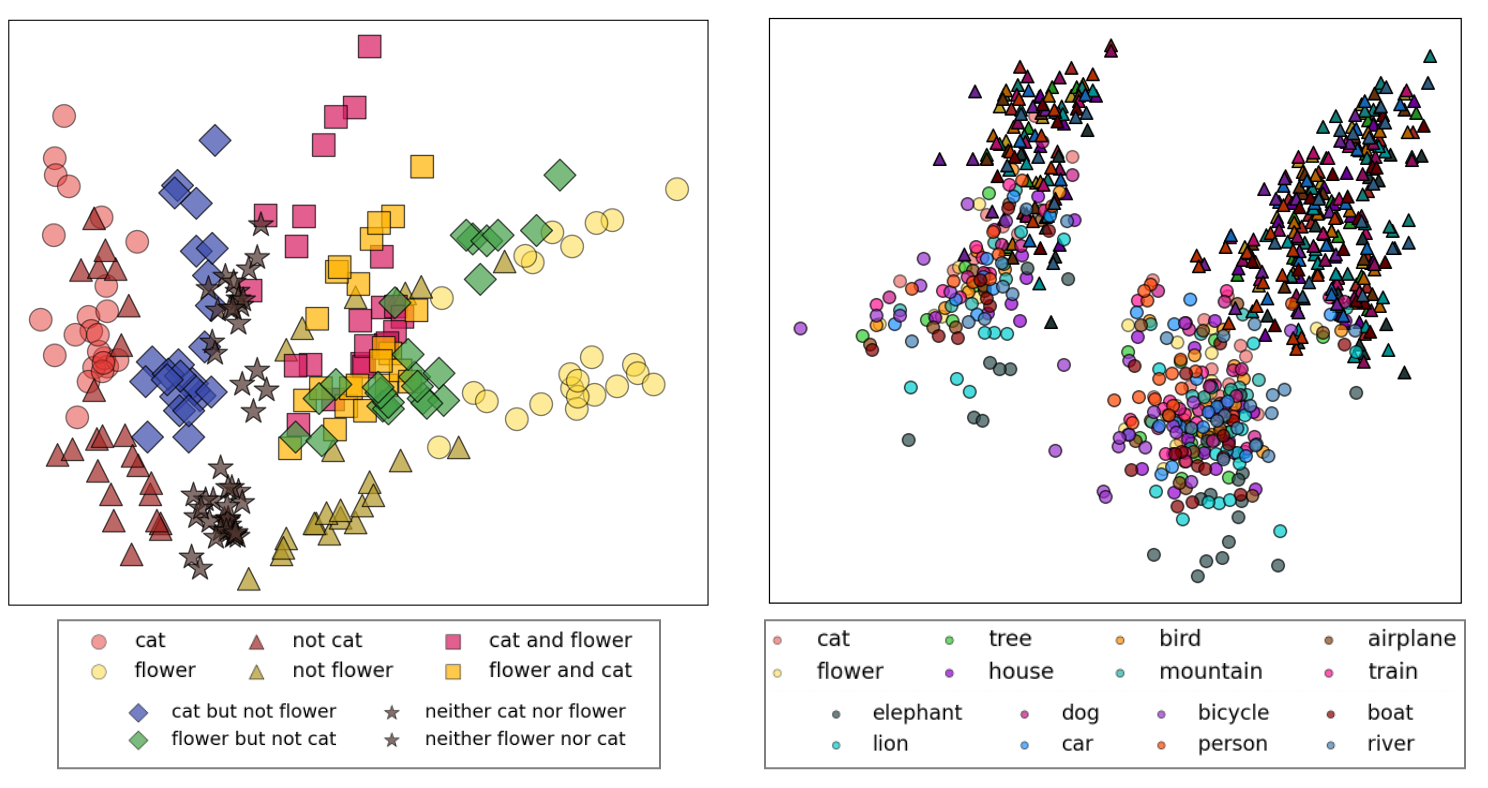}
\caption{PCA projections of caption embeddings for finetuned CLIP model on CC12M-NegCap. Each point represents a caption embedding. }
    \label{fig:ft_embeddings}
\end{figure}
\subsection{Template Selection Frequency}
\Cref{fig:template_frequency} analyzes how often different models select templates of type Affirmation, Negation, or Hybrid when answering multiple-choice questions—regardless of whether the selected answer is correct. This helps reveal systematic biases in model decision-making.

We observe that most CLIP-based models strongly overselect Negation templates, even when the correct answer is an Affirmation or Hybrid statement. This aligns with the results in \Cref{fig:mcq_breakdown}, where models struggle with Negation MCQs and tend to default to negated statements. This behavior supports our earlier claim of an \textit{affirmation bias}: models trained with CLIP-like objectives tend to ignore function words like ``not'' and collapse positive and negative statements in their embedding space.

\begin{table}[ht]
\centering
\caption{MCQ Total Accuracy (\%) across different datasets for various models}
\label{tab:mcq_accuracy}
\begin{tabular}{lccc}
\toprule
\textbf{Model} & \textbf{COCO} & \textbf{VOC2007} & \textbf{HardNeg-Syn} \\
\midrule
CLIP-OpenAI       & 16.27\% & 14.47\% & 18.24\% \\
CLIP-Laion400M    & 24.26\% & 27.01\% & {44.60\%} \\
CLIP-datacomp     & 19.73\% & 19.72\% & 34.10\% \\
NegCLIP           & 10.21\% & 8.51\%  & 17.03\% \\
ConCLIP           & 15.20\% & 20.43\% & 11.10\% \\
CLIP-L14          & 22.44\% & 23.69\% & 36.51\% \\
CLIP-H14          & {32.14\%} & {38.26\%} & 36.98\% \\
\bottomrule
\end{tabular}
\end{table}

\subsection{Template Embedding Analysis}

This subsection provides further details about the embedding analysis presented in \Cref{fig:combined_pca} of the main paper. We achieve this by:

\begin{enumerate}
    \item Specifying templates used to generate the embeddings.
    \item Expanding the embedding analysis to more models.
\end{enumerate}

To generate the embeddings for the PCA projections, we used five categories of templates: Affirmation (single object), Negation (single object), Affirmation (two objects), Hybrid (one object affirmed, one negated), and Double Negation (two objects negated). Each category contains 24 templates, except for Affirmation (two objects) which has 23. The templates vary sentence structure and wording while maintaining the same core meaning.

\begin{itemize}
    \item \textbf{Affirmation (single object):} 24 templates. Examples: "This image includes A", "A is present in this image", "This image shows A", "A is depicted in this image", "A appears in this image".
    \item \textbf{Negation (single object):} 24 templates. Examples: "This image does not include A", "A is not present in this image", "This image lacks A", "A is not depicted in this image", "A does not appear in this image".
    \item \textbf{Affirmation (two objects):} 23 templates. Examples: "This image includes A and B", "A and B are present in this image", "This image shows A and B", "A and B are depicted in this image", "A and B appear in this image".
    \item \textbf{Hybrid (one object affirmed, one negated):} 24 templates. Examples: "This image includes A but not B", "A is present in this image but not B", "This image shows A but not B", "This image features A but not B", "A appears in this image but not B".
    \item \textbf{Double Negation (two objects negated):} 24 templates. Examples: "This image includes neither A nor B", "Neither A nor B are present in this image", "This image shows neither A nor B", "Neither A nor B are depicted in this image", "Neither A nor B appear in this image".
\end{itemize}

While Figure 6 focused on CLIP, NegCLIP, and ConCLIP, \Cref{fig:all_embeddings} presents an additional visualization with PCA projections for other CLIP models (varying in size and pretraining datasets). This broader analysis will provide a more comprehensive view of how different CLIP models handle negation in the embedding space.

\section{Additional Insights and Context}

\vspace{4pt}
\noindent\textbf{D.1 How does this work fit into the broader landscape of negation and compositionality research?} \\
Prior benchmarks such as CREPE and CC-Neg introduced limited forms of negation in vision-language tasks, focusing on compositionality or constrained template-based generation. More recently, SPEC~\cite{peng2024synthesize} proposed fine-grained VQA tasks with a subset evaluating negation understanding. NaturalBench~\cite{linaturalbench} presents a vision-centric QA protocol that reveals large performance gaps between humans and top-tier VLMs (e.g., GPT-4o, Qwen2-VL), often caused by answer biases such as a tendancy to say “Yes.” over “No.”

Our work complements and extends these efforts with several contributions:
\begin{itemize}
  \item We introduce \textbf{NegBench}, a large-scale benchmark with 79K examples across retrieval and MCQ tasks, spanning images, video, and medical domains.
  \item We design \textbf{naturalistic negation prompts} using LLMs, covering a broad range of negation types and avoiding rigid linguistic templates.
  \item We generate \textbf{70M+ synthetic negation-enriched training samples}, supporting both contrastive and multiple-choice learning objectives.
  \item We conduct extensive experiments showing that our models \textbf{outperform prior negation-specific models (e.g., ConCLIP)} as well as SOTA VLMs (e.g., AIMv2) on negation tasks.
\end{itemize}

\vspace{4pt}
\noindent\textbf{D.2 What is the significance of model scaling experiments and comparisons to recent architectures like AIMv2?} \\
A common intuition is that larger models may better capture fine-grained distinctions such as negation. To evaluate this, we scale CLIP across ViT-B, L, and H variants, and additionally assess newer joint-embedding models such as SigLIP and AIMv2. Despite stronger performance on standard retrieval tasks, these models still struggle on MCQ-Neg and do not meaningfully close the gap—indicating that increased capacity alone does not resolve negation failures.

\vspace{4pt}
\noindent\textbf{D.3 How are negative object queries constructed in retrieval and MCQ settings?} \\
For datasets with dense annotations (COCO, VOC2007), we construct a co-occurrence matrix to identify object pairs that frequently appear together. We then generate negated prompts by selecting a plausible object that is \emph{absent} from the current image but typically co-occurs with present objects. This ensures that the negation is realistic and visually grounded, rather than relying on unlikely or artificially constructed distractors.

\vspace{4pt}
\noindent\textbf{D.4 What is the significance of the medical experiment, despite its simplicity?} \\
The medical retrieval experiment uses a simple binary decision setup, which offers a clean, interpretable upper bound on model capability. Models are tasked with distinguishing statements like “has pneumonia” versus “does not have pneumonia.” Despite the simplicity, we observe large performance drops under negation (up to 33\%) for domain-specialized VLMs such as BioMedCLIP and CONCH. This reveals a persistent failure mode with real-world clinical implications, where affirming or negating a condition must be handled with precision to avoid dire consequences.

\section{Dataset and Task Summary for \benchmark{}}
\label{sec:dataset_summary}

We provide a summary of the datasets and tasks used in \benchmark{}, a framework designed to evaluate Visual Language Models (VLMs) on their understanding of negation across different modalities, including images, videos, and medical imaging. The benchmark includes both retrieval and multiple-choice question (MCQ) tasks, with two variations: templated and LLM-paraphrased. For synthetic data, we generate 10,000 images using Stable Diffusion, which serve as hard negatives for one another, enabling a more focused evaluation of negation comprehension in text-to-image retrieval tasks.

Each dataset contributes to either Retrieval-Neg or MCQ-Neg tasks, except for CheXpert, which has two distinct tasks (Affirmation Control and Negation Understanding) in both MCQ and binary classification formats. Additionally, we utilize original retrieval captions for COCO (5,000) and MSR-VTT (1,000), expanding the overall dataset size. VOC2007 does not include a Retrieval-Neg task as it lacks retrieval-style captions.

The total number of task variations across all datasets in \benchmark{} is 18, and the total number of samples across all tasks and variations is 79,239. Table~\ref{tab:dataset_summary} summarizes the datasets, tasks, task versions, and sizes.

\begin{itemize}
    \item \textbf{COCO}: 5,000 retrieval captions and 5,914 MCQ questions, resulting in 10,000 retrieval problems and 11,828 MCQ problems with templated and LLM-paraphrased variations.
    \item \textbf{VOC2007}: 5,032 MCQ questions, leading to 10,064 total samples. No retrieval task is provided due to the absence of retrieval-style captions.
    \item \textbf{MSR-VTT}: 1,000 retrieval captions and 1,000 MCQ questions, resulting in 2,000 samples per task, including both variations.
    \item \textbf{CheXpert}: Two MCQ tasks (4-choice) and two binary classification tasks. The 4-choice MCQ covers 690 samples for affirmation and 1,587 for negation, while the binary tasks each include 690 samples.
    \item \textbf{HardNeg-Syn}: 10,000 synthetic images, used to create 20,000 retrieval and 20,000 MCQ problems across templated and LLM-paraphrased versions.
\end{itemize}

\begin{table*}[htbp]
\centering
\caption{\textbf{Summary of datasets and tasks in \benchmark{}.} Each task includes both templated and LLM-paraphrased versions, except for CheXpert tasks, which are templated only due to their straightforwardness (they directly evaluate diagnostic capabilities in the presence of negation words). The HardNeg-Syn dataset contains 10,000 synthetic images as hard negatives, offering a more targeted evaluation of negation understanding. The total number of task variations is 18, with a total of 79,239 samples across all tasks and variations.}
\label{tab:dataset_summary}
\renewcommand{\arraystretch}{1} 
\resizebox{\textwidth}{!}{
\begin{tabular}{llcccccl}
\toprule[1.5pt]
\textbf{Dataset} & \textbf{Task}             & \textbf{Templated} & \textbf{LLM-Paraphrased} & \textbf{Task Size} & \textbf{Notes} \\
\midrule[1.5pt]
\multirow{2}{*}{\textbf{COCO}} 
  & Retrieval-Neg                  & \checkmark & \checkmark & 10,000 & Image retrieval with negated captions. \\
  & MCQ-Neg                        & \checkmark & \checkmark & 11,828 & MCQ task with affirmative, negated, and hybrid options. \\
\midrule
\multirow{1}{*}{\textbf{VOC2007}}
  & MCQ-Neg                        & \checkmark & \checkmark & 10,064 & MCQ task. No Retrieval-Neg for VOC2007. \\
\midrule
\multirow{2}{*}{\textbf{MSR-VTT}} 
  & Retrieval-Neg                  & \checkmark & \checkmark & 2,000  & Video retrieval task with negated captions. \\
  & MCQ-Neg                        & \checkmark & \checkmark & 2,000  & Video-based MCQ task with temporal context. \\
\midrule
\multirow{2}{*}{\textbf{CheXpert} {(4-choice)}} 
  & Affirmation Control MCQ        & \checkmark & --         & 690    & Medical image MCQ with 4 choices. \\
  & Negation Understanding MCQ     & \checkmark & --         & 1,587  & MCQ task with negation. \\
\midrule
\multirow{2}{*}{\textbf{CheXpert} {(binary)}} 
  & Affirmation Control            & \checkmark & --         & 690    & Binary classification of medical images. \\
  & Negation Understanding         & \checkmark & --         & 690    & Binary classification, negated statements. \\
\midrule
\multirow{2}{*}{\textbf{HardNeg-Syn}} 
  & Retrieval-Neg                  & \checkmark & \checkmark & 20,000 & Synthetic image retrieval task. \\
  & MCQ-Neg                        & \checkmark & \checkmark & 20,000 & MCQ task for synthetic images with 4 answer choices. \\
\bottomrule[1.5pt]
\end{tabular}
}
\end{table*}

\vspace{100pt}
\subsection{Details of HardNeg-Syn Construction}
\label{sec:sd_details}

\setlength{\parskip}{0pt}  
\setlength{\itemsep}{0pt}  
\setlength{\topsep}{0pt}   
\begin{tcolorbox}[width=2\columnwidth, colback=gray!10, boxrule=0pt, sharp corners, colframe=gray!80, left=1mm, right=1mm, top=1mm, bottom=1mm]  
\vspace{-5pt}  
\subsubsection*{Object Label Selection}
We gather a wide range of object text labels from existing datasets like ImageNet.

\vspace{0pt}  
\subsubsection*{Scene Description}
For each selected object label (\textcolor{customgreen}{\bf\texttt{A}}), LLaMA 3.1 generates:\\
A \textcolor{customblue}{\bf\texttt{\{background\_description\}}} and a related object \textcolor{customred}{\bf\texttt{\{B\}}}, crafting realistic scene contexts.

\subsubsection*{Image Generation}
Using Stable Diffusion, we generate pairs of images:
\vspace{5pt}
\begin{adjustwidth}{2em}{1em}  
\textbf{Positive Image:} \textcolor{customblue}{\bf\texttt{\{background\_description\}}} with \textcolor{customgreen}{\bf\texttt{\{A\}}} next to \textcolor{customred}{\bf\texttt{\{B\}}}. \\[0.5em]
\textbf{Negative Image:} \textcolor{customblue}{\bf\texttt{\{background\_description\}}} with \textcolor{customgreen}{\bf\texttt{\{A\}}}, excluding \textcolor{customred}{\bf\texttt{\{B\}}} in the negative prompt to ensure its absence.
\end{adjustwidth}

\subsubsection*{Verification}
We use OWL-ViT~\citep{minderer2022simple} to verify the presence and absence of \textcolor{customgreen}{\bf\texttt{A}} and \textcolor{customred}{\bf\texttt{B}}.

\subsubsection*{Caption Generation}
Captions are generated using templates and paraphrased with LLaMA 3.1 for naturalness.
\vspace{-4pt}  
\end{tcolorbox}

\section{Visualizing the NegBench Evaluation Tasks}
\begin{figure*}[h]
    \centering
    \includegraphics[width=.8\linewidth]{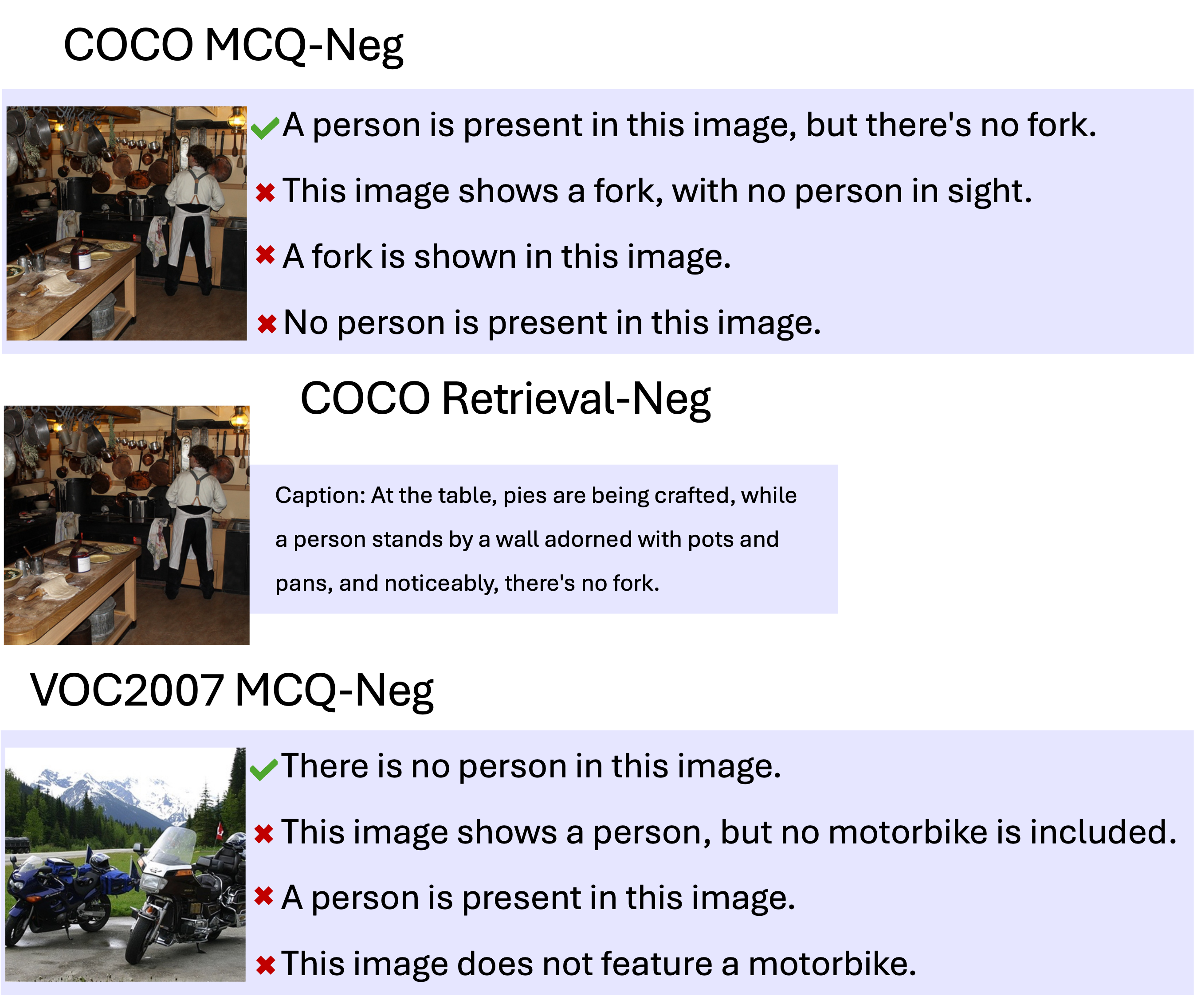}
    \caption{Examples of COCO and VOC2007 tasks, including Retrieval with negated captions and MCQ with negation.}
    \label{fig:coco_voc}
\end{figure*}

\begin{figure*}[h]
    \centering
    \includegraphics[width=.7\linewidth]{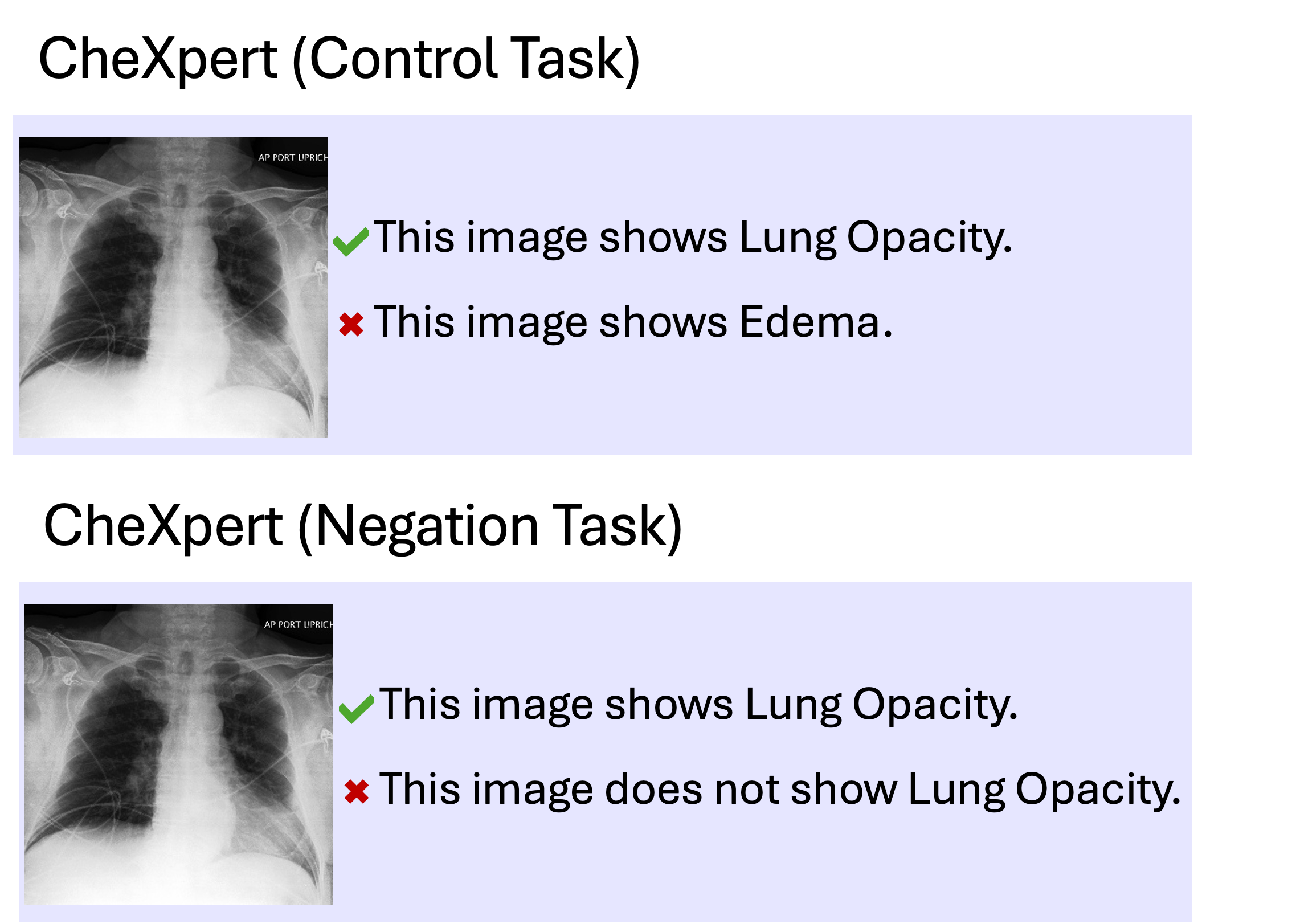}
    \caption{Examples of CheXpert MCQ tasks, including the Affirmation Control task and the Negation task.}
    \label{fig:chexpert_mcq}
\end{figure*}

\begin{figure*}[h]
    \centering
    \includegraphics[width=.9\linewidth]{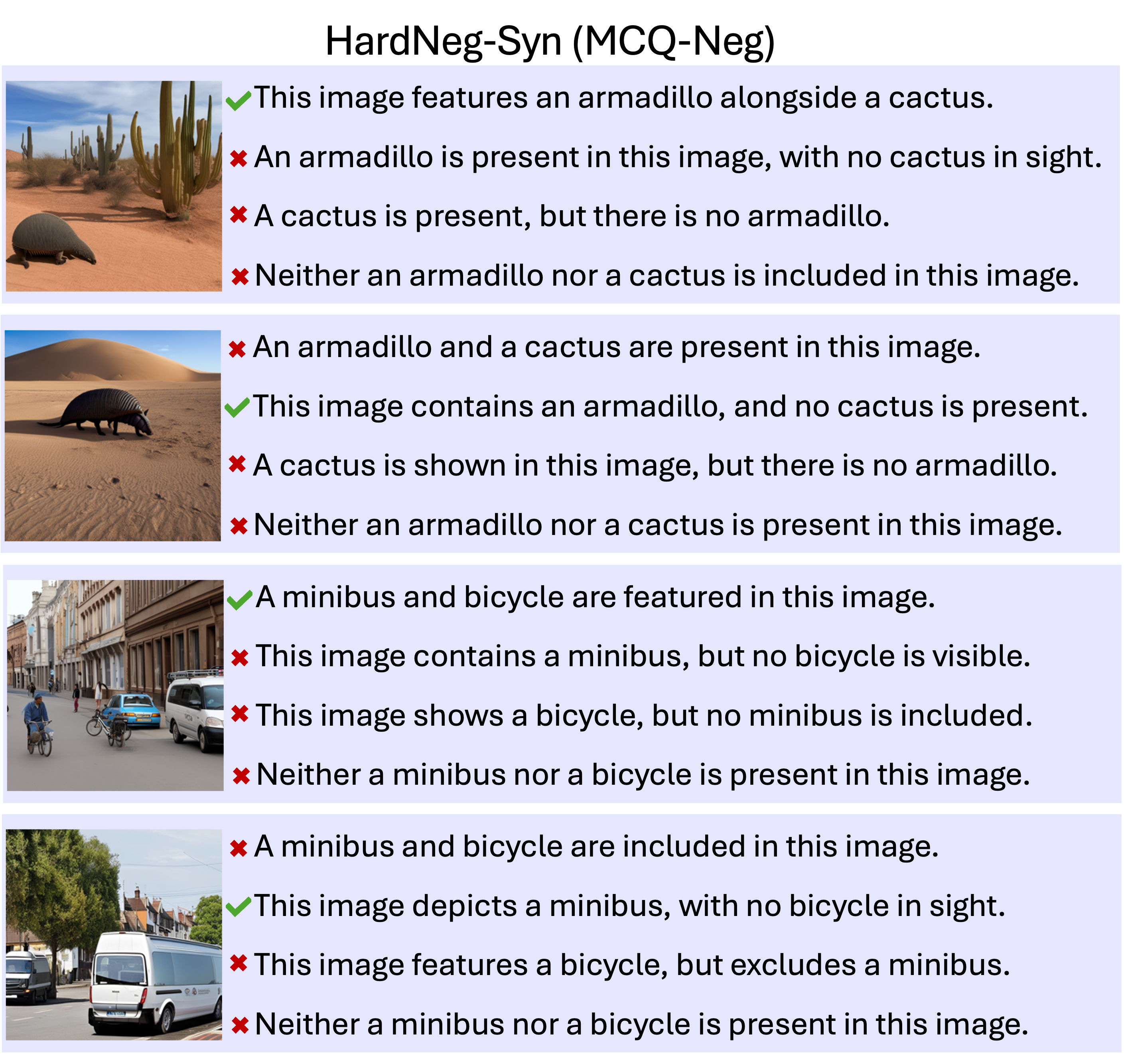}
    \caption{Examples of HardNeg-Syn (MCQ-Neg) tasks. Images in this dataset are constructed in pairs, with each pair differing by a single object (the cactus in the first pair), making the dataset particularly suitable for studying negation understanding.}
    \label{fig:hardneg_mcq}
\end{figure*}

\begin{figure*}[h]
    \centering
    \includegraphics[width=.9\linewidth]{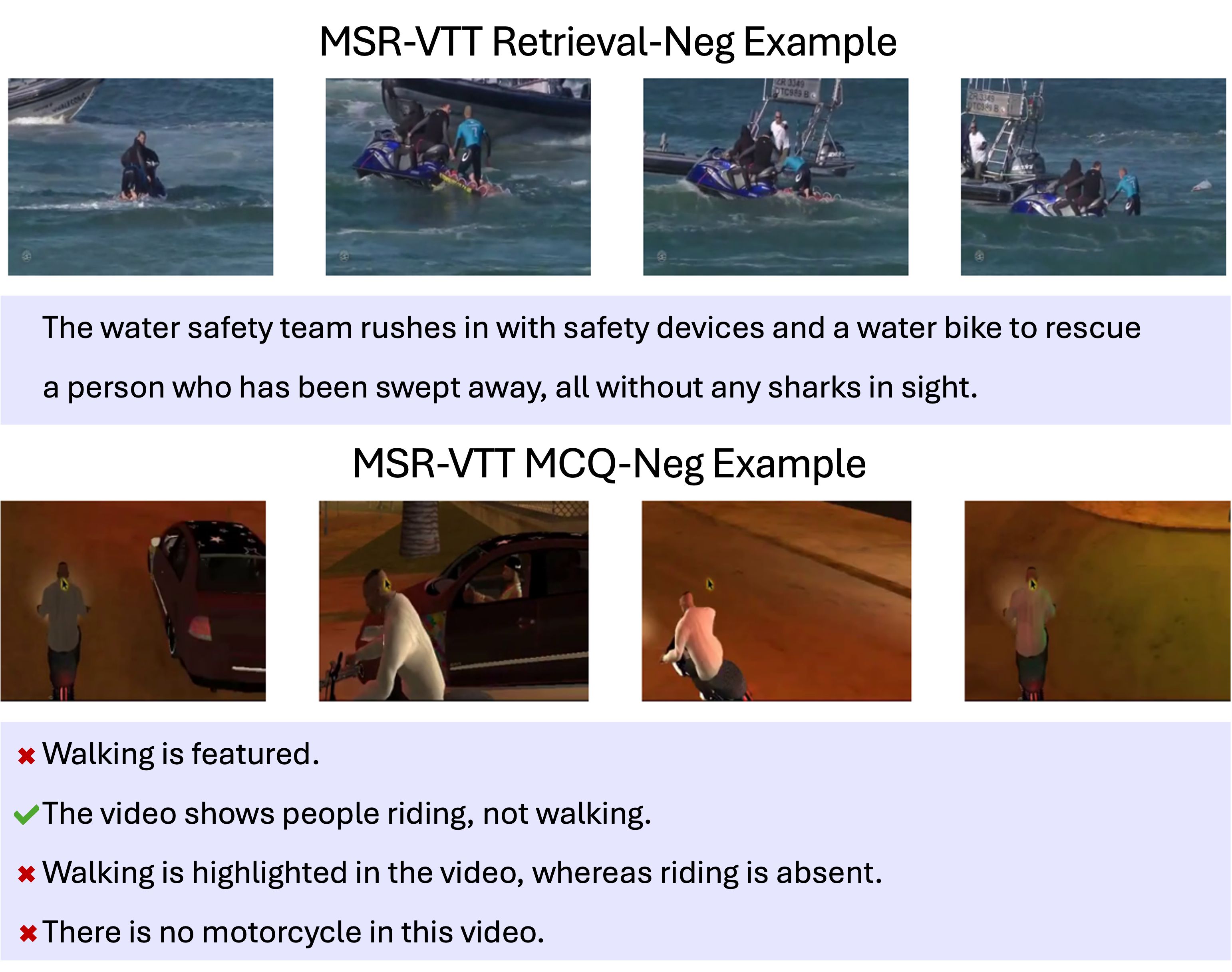}
    \caption{Examples of MSR-VTT tasks, including Retrieval-Neg (with negated captions about a complex water rescue scene) and MCQ-Neg (with answer choices about the presence or absence of actions like walking).}
    \label{fig:msrvtt}
\end{figure*}

In Figures \Cref{fig:coco_voc,fig:chexpert_mcq,fig:hardneg_mcq,fig:msrvtt}, we visualize a few samples from the NegBench retrieval and MCQ tasks we introduced in the paper. We note that the datasets are diverse in terms of the nature of visual domain and real-world applicability.

\end{document}